\documentstyle[12pt,psfig,named]{mitthesis}
\newcommand{\word}[1]{{\em #1}}
\newcounter{sentence}
\newenvironment{sentence}
{\begin{quote} \refstepcounter{sentence} (\thesentence) \hspace{1em}}
{\end{quote}}

\def\linkage#1#2#3{\vbox to 20bp{\vss\special{"#1 #2 #3 diagram}}}

\newcommand{\myvspace}[1]{{\smallskip \parindent=0pt \vspace*{#1}\smallskip}}

\newcommand{\mylinkage}[2] {
\vbox to 20bp{\vss\special{"#1 #2 mydiagram}}
\vspace{11pt}
}

\newlength{\toppush}
\newcounter{problemsetnum}

\newcommand{\examtitle}[2]
{\noindent\vspace*{-\toppush}\newline\parbox{6.5in}
{Massachusetts Institute of Technology\newline
6.851J/18.414J: Theory of Algorithms\hfill#1\newline
Professor Ronald L. Rivest\hfill#2\vspace*{-1ex}\newline
\mbox{}\hrulefill\mbox{}}\vspace*{1ex}\mbox{}\newline
\begin{center}{\Large\bf #1}\end{center}}

\makeatletter
\newcommand{\exambooklet}[2]{
 \thispagestyle{empty}
 \markboth{6.851J/18.414J #1}{6.851J/18.414J #1}
 \pagestyle{myheadings}\examtitle{#1}{#2}
 \renewcommand{\theproblem}{Problem \arabic{problemnum}}
 \renewcommand{\problem}{\newpage
 \item \let\@currentlabel=\theproblem
 \markboth{6.851J/18.414J #1, \theproblem}{6.851J/18.414J #1, \theproblem}}
}
\makeatother

\newcounter{problemnum}
\newcommand{\theproblem}{Problem \theproblemsetnum-\arabic{problemnum}}
\newenvironment{problems}{
	\begin{list}{{\bf \theproblem. \hspace*{0.5em}}}
	{\setlength{\leftmargin}{0em}
	 \setlength{\rightmargin}{0em}
	 \setlength{\labelwidth}{0em}
	 \usecounter{problemnum}}}{\end{list}}
\makeatletter
\newcommand{\problem}{\item \let\@currentlabel=\theproblem}
\makeatother

\newcounter{problempartnum}[problemnum]
\newenvironment{problemparts}{
	\begin{list}{{\bf (\alph{problempartnum})}}
	{\setlength{\leftmargin}{2.5em}
	 \setlength{\rightmargin}{2.5em}
	 \setlength{\labelsep}{0.5em}}}{\end{list}}

\newcounter{exercisenum}
\newcommand{\theexercise}{Exercise \theproblemsetnum-\arabic{exercisenum}}

\makeatletter
\newcommand{\exercise}{\item \let\@currentlabel=\theexercise}
\makeatother

\newcounter{exercisepartnum}[exercisenum]

\makeatletter
\def\fnum@figure{{\bf Figure \thefigure}}
\def\fnum@table{{\bf Table \thetable}}
\makeatother

\makeatletter
\newif\ifcodeinbox
\newif\ifdoubledigit
\newif\ifli

\newcounter{codelinenumber}
\newcommand{\zeroli}{\setcounter{codelinenumber}{0}}

\def\@startline{\global\@curtabmar\@nxttabmar\relax
   \global\@curtab\@curtabmar\setbox\@curline\hbox
    {}\@startfield\global\lifalse\strut}

\newenvironment{code}{\global\codeinboxtrue%
\setbox\strutbox\hbox{\vrule height 9pt depth 4pt width0pt}%
\noindent\begin{tabbing}%
\zeroli\setlength{\tabbingsep}{1em}
\hspace*{1em}\=999\ifdoubledigit9\fi
\=\ {\bf if} \={\bf then} \={\bf if} \={\bf then}
        \={\bf if} \={\bf then} \={\bf if} \={\bf then} \={\bf if} \={\bf then}
        \={\bf if} \=\+\+\kill}{\end{tabbing}\global\codeinboxfalse}

\newcommand{\codebox}[1]{
  \setbox0=\vbox{\begin{code}#1\end{code}}
  \ifnum\c@codelinenumber>9
    \global\doubledigittrue
  \else
    \doubledigitfalse
  \fi
  \vskip1sp\noindent\hskip-14pt
  \parbox{\textwidth}{\begin{code}\protect#1\end{code}}}

\newcommand{\li}{\global\litrue\stepcounter{codelinenumber}
  \ifdoubledigit
    \hbox to8pt{\hss\thecodelinenumber\hskip5pt}
  \else
    \hbox to8pt{\hskip-1pt\thecodelinenumber\hss}
  \fi
  \xdef\@currentlabel{\p@codelinenumber\thecodelinenumber}\'}

\makeatother

\newcommand{\For}{{\bf for} }
\newcommand{\To}{{\bf to} }

\newcommand{\Downto}{{\bf downto} }

\newcommand{\If}{{\bf if} }
\newcommand{\Then}{{\bf then} }

\makeatletter
\def\hanglistlabel#1{\hspace\labelsep}
\def\hanglist{\leftmarginii=1pc
\list{}{\labelsep=0pt\labelwidth\z@ \itemindent-1pc
\let\makelabel\hanglistlabel}}

\makeatother

\load{\normalsize}{\sc}
\load{\small}{\sc}
\newcommand{\proc}[1]
  {\ifmmode\mathord{\mathcode`-="702D\sc#1\mathcode`\-="2200}\else{\sc#1}\fi}
\newcommand{\id}[1]
  {\ifmmode\mathord{\mathcode`-="702D\it#1\mathcode`\-="2200}\else$\mathord{\mathcode`-="702D\it#1\mathcode`\-="2200}$\fi}
\newcommand{\lang}[1]
  {\ifmmode\mathord{\mathcode`-="702D\rm#1\mathcode`\-="2200}\else{\rm#1}\fi}
\newcommand{\const}[1]{\ifmmode\mathord{\mathcode`-="702D\sc #1\mathcode`\-="2200}\else$\mathord{\mathcode`-="702D\sc #1\mathcode`\-="2200}$\fi}

\newcommand{\ang}[1]{
  \ifmmode{\left\langle #1 \right\rangle}      
  \else{$\left\langle${#1}$\right\rangle$}\fi} 

\renewcommand{\choose}[2]{{{#1}\atopwithdelims(){#2}}}

\newtheorem{theorem}{Theorem}

\newtheorem{xample}{Example}

\newcommand{\startproof}{\noindent{\bf Proof:}}
\def\squarebox#1{\hbox to #1{\hfill\vbox to #1{\vfill}}}
\newcommand{\qedbox}{\vbox{\hrule\hbox{\vrule\squarebox{.667em}\vrule}\hrule}}
\newcommand{\qed}{\nopagebreak\mbox{}\hfill\qedbox\smallskip}
\newenvironment{proof}{\startproof}{\qed}

\begin{document}

\title{Discovery of Linguistic Relations Using Lexical Attraction}

\author{Deniz Yuret}
\department{Department of Electrical Engineering and Computer Science}
\degree{Doctor of Philosophy}
\degreemonth{May}
\degreeyear{1998}
\thesisdate{May 15, 1998}

\supervisor{Patrick H. Winston}
{Ford Professor of Artificial Intelligence and Computer Science}

\chairman{Arthur C. Smith}{Chairman, Department Committee on Graduate Students}

\maketitle

\newpage
\setcounter{savepage}{\thepage}
\begin{abstractpage}

This work has been motivated by two long term goals: to understand how
humans learn language and to build programs that can understand
language.  Using a representation that makes the relevant features
explicit is a prerequisite for successful learning and understanding.
Therefore, I chose to represent relations between individual words
explicitly in my model.  {\em Lexical attraction} is defined as the
likelihood of such relations.  I introduce a new class of
probabilistic language models named {\em lexical attraction models}
which can represent long distance relations between words and I
formalize this new class of models using information theory.

Within the framework of lexical attraction, I developed an
unsupervised language acquisition program that learns to identify
linguistic relations in a given sentence.  The only explicitly
represented linguistic knowledge in the program is lexical attraction.
There is no initial grammar or lexicon built in and the only input is
raw text.  Learning and processing are interdigitated.  The processor
uses the regularities detected by the learner to impose structure on
the input.  This structure enables the learner to detect higher level
regularities.  Using this bootstrapping procedure, the program was
trained on 100 million words of Associated Press material and was able
to achieve 60\% precision and 50\% recall in finding relations between
content-words.  Using knowledge of lexical attraction, the program can
identify the correct relations in syntactically ambiguous sentences
such as ``I saw the Statue of Liberty flying over New York.''

\end{abstractpage}

\newpage

\section*{Acknowledgments}

I am grateful to Carl de Marcken for his valuable insights, Alkan
Kabak\c{c}{\i}o\u{g}lu for sharing my interest in math, and Ayla
O\u{g}u\c{s} for her endless patience.  I am thankful to my mother and
father for the importance they placed on my education.  I am indebted
to my advisors Patrick Winston for teaching me AI, Boris Katz for
teaching me language, and Marvin Minsky for teaching me to avoid bad
ideas.

\pagestyle{plain}
\tableofcontents
\listoffigures

\chapter{Language Understanding and Acquisition}
\label{intro}

This work has been motivated by a desire to explain language learning
on one hand and to build programs that can understand language on the
other.  I believe these two goals are very much intertwined.  As with
many other areas of human intelligence, language proved not to be
amenable to small models and simple rule systems.  Unlocking the
secrets of learning language from raw data will open up the path to
robust natural language understanding.

I believe what makes humans good learners is not sophisticated
learning algorithms but having the right representations.  Evolution
has provided us with cognitive transducers that make the relevant
features of the input explicit.  The representational primitives for
language seems to be the linguistic relations like subject-verb,
verb-object.  The standard phrase-structure formalism only indirectly
represents such relations as side-effects of the constituent-grouping
process.  I adopted a formalism which takes relations between
individual words as basic primitives.  Lexical attraction gives the
likelihood of such relations.  I built a language program in which the
only explicitly represented linguistic knowledge is lexical
attraction.  It has no grammar or a lexicon with parts of speech.

My program does not have different stages of learning and processing.
It learns while processing and gets better as it is presented with
more input.  This makes it possible to have a feedback loop between
the learner and the processor.  The regularities detected by the
learner enable the processor to assign structure to the input.  The
structure assigned to the input enables the learner to detect higher
level regularities.  Starting with no initial knowledge, and seeing
only raw text input, the program is able to bootstrap its acquisition
and show significant improvement in identifying meaningful relations
between words.

The first section presents lexical attraction knowledge as a solution
to the problems of language acquisition and syntactic disambiguation.
The second section describes the bootstrapping procedure in more
detail.  The third section presents snapshots from the learning
process.  Chapter~\ref{demo} gives more examples of learning.
Chapter~\ref{theory} explains the computational, mathematical and
linguistic foundations of the lexical attraction models.
Chapter~\ref{learning} describes the program and its results in more
detail.  Chapter~\ref{contributions} summarizes the contributions of
this work.

\section{The case for lexical attraction}

Lexical attraction is the measure of affinity between words, i.e. the
likelihood that two words will be related in a given sentence.
Chapter~\ref{theory} gives a more formal definition.  The main premise
of this thesis is that knowledge of lexical attraction is central to
both language understanding and acquisition.  The questions addressed
in this thesis are how to formalize, acquire and use the lexical
attraction knowledge.  This section argues that language acquisition
and syntactic disambiguation are similar problems, and knowledge of
lexical attraction is a powerful tool that can be used to solve both
of them.

\subsubsection*{Language understanding}

Syntax and semantics play complementary roles in language
understanding.  In order to understand language one needs to identify
the relations between the words in a given sentence.  In some cases,
these relations may be obvious from the meanings of the words.  In
others, the syntactic markers and the relative positions of the words
may provide the necessary information.  Consider the following
examples:

\begin{sentence} \label{sLenat}
I saw the Statue of Liberty flying over New York.
\end{sentence}

\begin{sentence} \label{sSchank}
I hit the boy with the girl with long hair with a hammer with
vengeance.
\end{sentence}

In sentence (\ref{sLenat}) either the subject or the object may be
doing the flying.  The common interpretation is that I saw the Statue
of Liberty while I was flying over New York.  If the sentence was {\em
``I saw the airplane flying over New York''}, most people would
attribute flying to the \word{airplane} instead.  The two sentences
are syntactically similar but the decision can be made based on which
words are more likely to be related.

Sentence (\ref{sSchank}) ends with four prepositional phrases.  Each
of these phrases can potentially modify the subject, the verb, the
object, or the noun of a previous prepositional phrase, subject to
certain constraints discussed in Chapter~\ref{theory}.  In other
words, syntax leaves the question of which words are related in this
sentence mostly open.  The reader decides based on the likelihood of
potential relations.

\begin{sentence} \label{sChomsky}
Colorless green ideas sleep furiously.
\end{sentence}

In contrast, sentence (\ref{sChomsky}) is a classical example used to
illustrate the independence of grammaticality from
meaningfulness\footnote{Sentence (\ref{sChomsky}) is from Chomsky
\cite{Chomsky57}.  Sentence (\ref{sLenat}) is attributed to Lenat.
Sentence (\ref{sSchank}) is from Schank \cite{Schank73}.}.  Even
though none of the words in this sentence go together in a meaningful
way, we can nevertheless tell their relations from syntactic clues.

These examples illustrate that syntax and semantics independently
constrain the possible interpretations of a sentence.  Even though
there are cases where either syntax or semantics alone is enough to
get a unique interpretation, in general we need both.  What we need
from semantics in particular is the likelihood of various relations
between words.

\subsubsection*{Language acquisition}

Children start mapping words to concepts before they have a full grasp
of syntax.  At that stage, the problem facing the child is not unlike
the disambiguation problem in sentences like (\ref{sLenat}) and
(\ref{sSchank}).  In both cases, the listener is trying to identify
the relations between the words in a sentence and syntax does not
help.  In the case of the child, syntactic rules are not yet known.
In the case of the ambiguous sentences, syntactic rules cannot
differentiate between various possible interpretations.

Similar problems call for similar solutions.  Just as we are able to
interpret ambiguous sentences relying on the likelihood of potential
relations, the child can interpret a sentence with unknown syntax the
same way.

\begin{figure}[h]
\centering
\fbox{\psfig{file=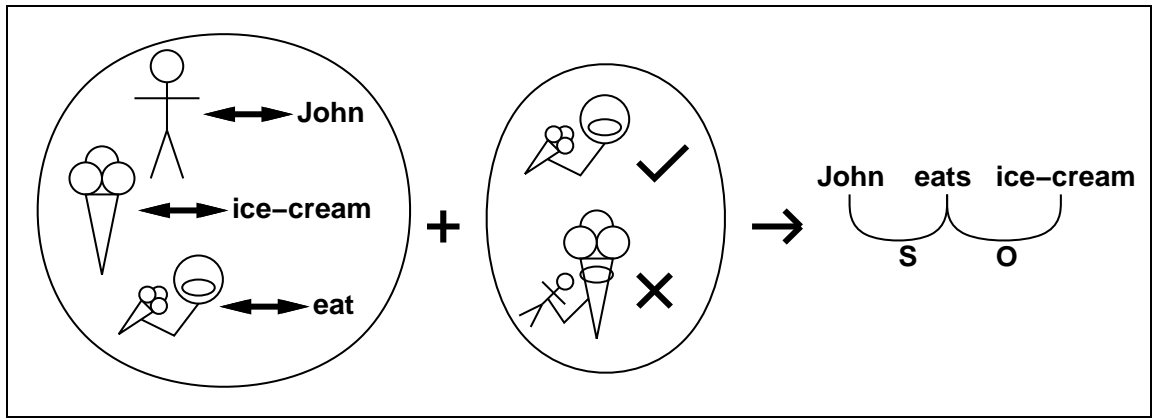,width=5.9in}}
\caption{Semantic judgments help bootstrap syntax.}
\label{ice-cream}
\end{figure}

Figure~\ref{ice-cream} illustrates this language acquisition path.
Exposure to language input teaches the child which words map to which
concepts.  Experience with the world teaches him the likelihood of
certain relations between concepts.  With this knowledge, it becomes
possible to identify certain linguistic relations in a sentence before
a complete syntactic analysis is possible.

With the pre-syntax identification of linguistic relations, syntactic
acquisition can be bootstrapped.  In the sentence ``John eats
ice-cream'', John is the subject of eating and ice-cream is the
object.  English relies on the SVO word order to identify these roles.
Other languages may have different word ordering or use other
syntactic markers.  Once the child identifies the subject and the
object semantically, he may be able to learn what syntactic rule his
particular language uses.  Later, using such syntactic rules, the
child can identify less obvious relations as in sentence
(\ref{sChomsky}) or guess the meanings of unknown words based on their
syntactic role.

In language acquisition, as in disambiguation, knowing how likely two
words are related is of central importance.  This knowledge is
formalized with the concept of {\em lexical attraction}.

\section{Bootstrapping acquisition}

Learning and encoding world experience with computers has turned out
to be a challenging problem.  Current common sense reasoning systems
are still in primitive stages.  This suggests the alternative of using
large corpora to gather information about the likelihood of certain
relations between words.

However, using large corpora presents the following chicken-and-egg
problem.  In order to gather information about the likelihood that two
words will be related, one first has to be able to detect that they
are related.  But this requires knowing syntax, which is what we were
trying to learn in the first place.

\begin{figure}[h]
\centering
\fbox{\psfig{file=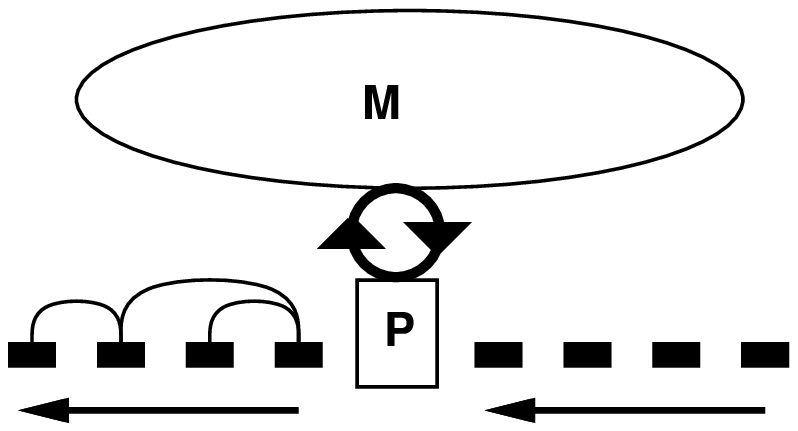,height=2in}}
\caption{Interdigitating learning and processing to bootstrap acquisition.}
\label{bootstrap}
\end{figure}

To get out of this loop, the learning program needs a bootstrapping
mechanism.  The key to bootstrapping lies in interdigitating learning
and processing.  Figure~\ref{bootstrap} illustrates this feedback loop.
With no initial knowledge of syntax, the processor P starts making
inaccurate analyses and memory M starts building crude lexical
attraction knowledge based on them.  This knowledge eventually helps
the processor detect relations more accurately, which results in
better quality lexical attraction knowledge in the memory.

Based on this idea, I built a language learning program that
bootstraps with no initial knowledge, reads examples of free text, and
learns to discover linguistic relations that can form a basis for
language understanding.

\begin{figure}[h]
\centering
\mbox{\psfig{file=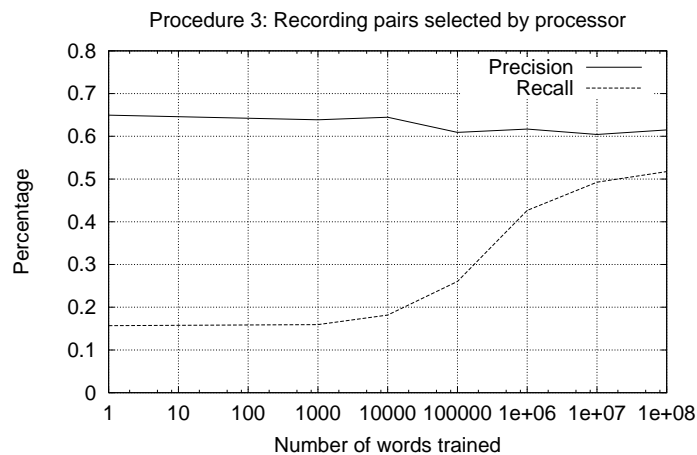,height=2.5in}}
\caption{Accuracy in relations between content-words}
\label{intro-result}
\end{figure}

The program was evaluated using its accuracy in relations between
content-words, e.g. nouns, verbs, adjectives and adverbs.  The
accuracy was measured using precision and recall.  The precision is
defined as the percentage of relations found by the program that were
correct.  The recall is defined as the percentage of correct relations
that were found by the program.  The program was able to achieve 60\%
precision and 50\% recall.  Previous work in unsupervised language
acquisition showed little improvement when started with zero
knowledge.  Figure~\ref{intro-result} shows the improvement my program
shows.  Detailed results are given in Chapter~\ref{learning}.

\section{Learning to process a simple sentence}

\begin{figure}[ht]
\fbox{
\parbox{5.8in}{
$N=0$

\linkage
{[(*)(these)(people)(also)(want)(more)(government)(money)(for)(education)( . )( * )]}
{[]}
{[0 ]}
\smallskip

$N=1,000$

\myvspace{5pt}
\linkage
{[(*)(these)(people)(also)(want)(more)(government)(money)(for)(education)( . )( * )]}
{[[0 1 0 ()][10 11 0 ()]]}
{[0 ]}
\smallskip

$N=10,000$

\myvspace{35pt}
\linkage
{[(*)(these)(people)(also)(want)(more)(government)(money)(for)(education)( . )( * )]}
{[[0 1 0 ()][0 2 1 ()][0 3 2 ()][0 4 3 ()][7 8 0 ()][10 11 0 ()]]}
{[0 ]}
\smallskip

$N=100,000$

\myvspace{20pt}
\linkage
{[(*)(these)(people)(also)(want)(more)(government)(money)(for)(education)( . )( * )]}
{[[0 1 0 ()][1 2 0 ()][3 4 0 ()][2 4 1 ()][2 5 2 ()][7 8 0 ()][6 8 1 ()][5 8 2 ()][8 9 0 ()][9 10 0 ()][10 11 0 ()]]}
{[0 ]}
\smallskip

$N=1,000,000$ 

\myvspace{20pt}
\linkage
{[(*)(these)(people)(also)(want)(more)(government)(money)(for)(education)( . )( * )]}
{[[0 1 0 ()][1 2 0 ()][3 4 0 ()][2 4 1 ()][6 7 0 ()][5 7 1 ()][4 7 2 ()][7 8 0 ()][7 9 1 ()][9 10 0 ()][10 11 0 ()]]}
{[0 ]}
}}
\caption{Discovering relations in a simple sentence.}
\label{simple}
\end{figure}

Figure~\ref{simple} shows how the program gradually discovers the
correct relations in a simple sentence.  $N$ denotes the number of
words used for training.  All words are lowercased.  The symbol *
marks the beginning and the end of the sentence.  The links are
undirected.  In Chapter~\ref{theory}, I show that the directions
of the links are immaterial for the training process.

Before training ($N=0$) the program has no information and no links
are found.  At 1,000 words the program has discovered that a period
usually ends the sentence and the word \word{these} frequently starts
one.  At 10,000 words, not much has changed.  The frequent collocation
\word{money for} is discovered.  More words link to the left *
marker.  Notice that \word{want}, for example, almost never starts a
sentence.  It is linked to the left * marker because as more links
are formed, the program is able to see longer distance correlations.

The lack of meaningful links up to this point can be explained by the
nature of word frequencies.  A typical word in English has a frequency
in the range of $1/10,000$ to $1/50,000$.  A good word frequency
formula based on Zipf's law is $\frac{1}{10n}$ where $n$ is the rank
of the word \cite{Zipf49,Shannon51}.  This means that after 10,000
words of training, the program has seen most words only once or twice,
not enough to determine their correlations.

At 100,000 words, the program discovers more interesting links.  The
word \word{people} is related to \word{want}, \word{these} modifies
\word{people}, and \word{also} modifies \word{want}.  The link between
\word{more} and \word{for} is a result of having seen many instances of
\word{more X for Y}.

The reason for many links to the word \word{for} at $N=100,000$
deserves some explanation.  We can separate all English words into two
rough classes called function words and content words.  Function words
include closed class words, usually of grammatical function, such as
prepositions, conjunctions, and auxiliary verbs.  Content words
include words bearing actual semantic content, such as nouns, verbs,
adjectives, and adverbs.  Function words are typically much more
frequent.  The most frequent function word \word{the} is seen $5\%$ of
the time, others typically are in the $1/100$ to $1/1,000$ range.
This means that the program first discovers
function-word-function-word links, like the one between period and *.
Next, the function-word-content-word links are discovered, like the
ones connecting \word{for}.  The content-word-content-word links are
discovered much later.

After 1,000,000 words of training, the program is able to discover the
correct links for this sentence.  The verb is connected to the subject
and the object.  The modifiers are connected to their heads.  The
words \word{money} and \word{education} related by the preposition
\word{for} are linked together.

\chapter{Discovery of Linguistic Relations: A Demonstration}
\label{demo}

This chapter presents snapshots from the learning process.  The
underlying theory and the algorithm follows in the next chapters.  The
examples in this chapter were chosen to illustrate the handling of
various linguistic phenomena.  Formal performance results and a
critical evaluation of the program's shortcomings are given in
Chapter~\ref{learning}.

\begin{figure}[h]
\centering\fbox{\psfig{file=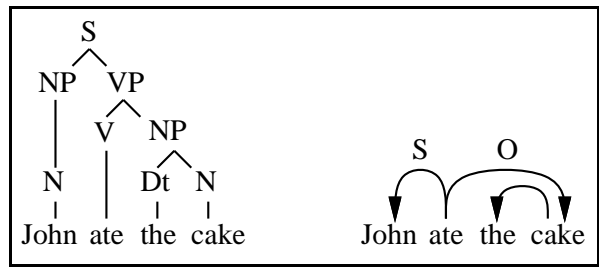}}
\caption{Phrase structure versus dependency structure.}
\label{dependency}
\end{figure}

The syntax is represented in a dependency formalism.
Figure~\ref{dependency} contrasts the phrase structure and the
dependency representations of a sentence.  The phrase structure
representation is based on forming higher order units by combining
words or phrases.  The dependency representation is based on explicit
representation of the relationships between individual words.
Chapter~\ref{theory} gives a more formal definition.

For the examples in this chapter, the program was trained on a corpus
of Associated Press Newswire material\footnote{AP Newswire 1988-1990
data from the TIPSTER Information Retrieval Text Research
Collection.}.  It was stopped at various points during training and
given the example sentences for processing.

\section{Long-distance links}

\begin{figure}[h]
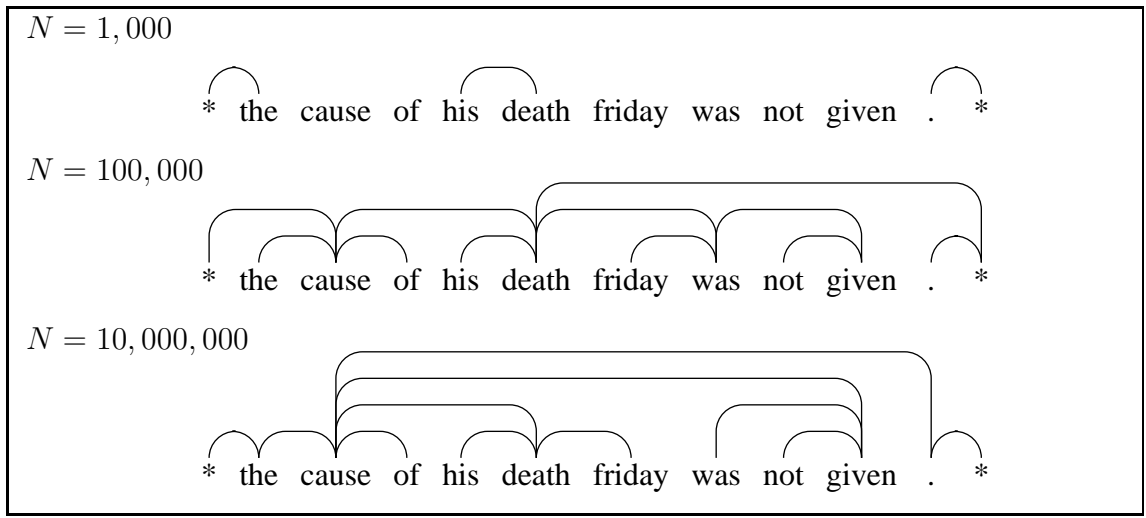

\centering
\fbox{
\parbox{5.8in}{
$N=1,000$

\myvspace{10pt}
\linkage
{[(*)(the)(cause)(of)(his)(death)(friday)(was)(not)(given)( . )( * )]}
{[[0 1 0 ()][4 5 0 ()][10 11 0 ()]]}
{[0 ]}

$N=100,000$

\myvspace{20pt}
\linkage
{[(*)(the)(cause)(of)(his)(death)(friday)(was)(not)(given)( . )( * )]}
{[[1 2 0 ()][0 2 1 ()][2 3 0 ()][4 5 0 ()][2 5 1 ()][6 7 0 ()][5 7 1 ()][8 9 0 ()][7 9 1 ()][10 11 0 ()][5 11 2 ()]]}
{[0 ]}

$N=10,000,000$

\myvspace{30pt}
\linkage
{[(*)(the)(cause)(of)(his)(death)(friday)(was)(not)(given)( . )( * )]}
{[[0 1 0 ()][1 2 0 ()][2 3 0 ()][4 5 0 ()][2 5 1 ()][5 6 0 ()][8 9 0 ()][7 9 1 ()][2 9 2 ()][2 10 3 ()][10 11 0 ()]]}
{[0 ]}
}}
\caption{Discovering long distance relations.}
\label{long}
\end{figure}

Most of the links in Figure~\ref{simple} spanned a few words.
Figure~\ref{long} shows that the program is also capable of handling
longer distance relations.  The sentence has a long noun phrase headed
by the noun \word{cause}.  It is this \word{cause} which is not
\word{given}, a link that spans the length of the sentence.

At 1,000 words, you see again that nothing much interesting is
discovered.  At 100,000 words, the program is able to relate the
\word{cause} to the \word{death} but longer distance relations are
still missing.  After ten million words of training, the attraction
between the word \word{cause} and the word \word{given} is discovered
and the correct link is created.

\newpage
\section{Complex noun phrase}

\begin{figure}[h]
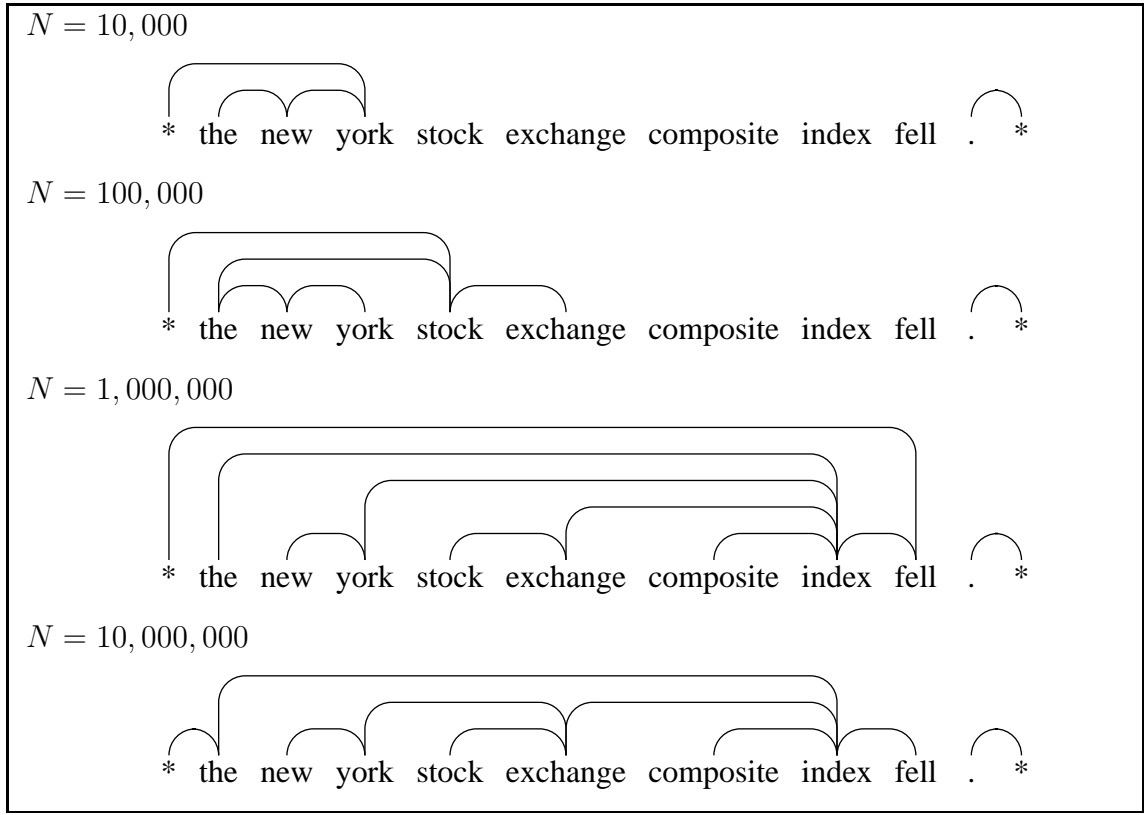

\centering
\fbox{
\parbox{5.8in}{
$N=10,000$

\myvspace{20pt}
\linkage
{[(*)(the)(new)(york)(stock)(exchange)(composite)(index)(fell)( . )( * )]}
{[[1 2 0 ()][2 3 0 ()][0 3 1 ()][9 10 0 ()]]}
{[0 ]}

$N=100,000$

\myvspace{30pt}
\linkage
{[(*)(the)(new)(york)(stock)(exchange)(composite)(index)(fell)( . )( * )]}
{[[1 2 0 ()][2 3 0 ()][1 4 1 ()][0 4 2 ()][4 5 0 ()][9 10 0 ()]]}
{[0 ]}

$N=1,000,000$

\myvspace{50pt}
\linkage
{[(*)(the)(new)(york)(stock)(exchange)(composite)(index)(fell)( . )( * )]}
{[[2 3 0 ()][4 5 0 ()][6 7 0 ()][5 7 1 ()][3 7 2 ()][1 7 3 ()][7 8 0 ()][0 8 4 ()][9 10 0 ()]]}
{[0 ]}

$N=10,000,000$

\myvspace{30pt}
\linkage
{[(*)(the)(new)(york)(stock)(exchange)(composite)(index)(fell)( . )( * )]}
{[[0 1 0 ()][2 3 0 ()][4 5 0 ()][3 5 1 ()][6 7 0 ()][5 7 1 ()][1 7 2 ()][7 8 0 ()][9 10 0 ()]]}
{[0 ]}
}}
\caption{Structure of a complex noun phrase.}
\label{nyse}
\end{figure}

One of the most difficult problems for non-lexicalized language
systems is to analyze the structure of a complex noun phrase.  The
noun phrase {\em ``the New York Stock Exchange Composite Index''} in
Figure~\ref{nyse} turns into {\em ``determiner adjective noun noun
noun adjective noun''} when seen as just parts of speech.  The parts
of speech do not give enough information to assign a meaningful
structure to the phrase.

My program collects information about individual words but it has no
concept of parts of speech.  It is able to discover the structure of
the complex noun phrase in Figure~\ref{nyse} because pieces of that
noun phrase are repetitively used elsewhere.

At 10,000 words, it discovers the group {\em ``new york''}.  At
100,000 words, it discovers {\em ``stock exchange''}.  At a million
words it discovers {\em ``composite index''}.  And finally at ten
million words it figures out the correct relations between these
pieces.

\section{Syntactic ambiguity}

I have argued in Chapter~\ref{intro} that we need semantic judgments
to interpret syntactically ambiguous sentences.  Specifically what we
need is information about the likelihood of various relations between
words, i.e. lexical attraction information.  This section presents
several examples of syntactic ambiguity and demonstrate how lexical
attraction information helps to resolve the ambiguity.

\begin{figure}[h]
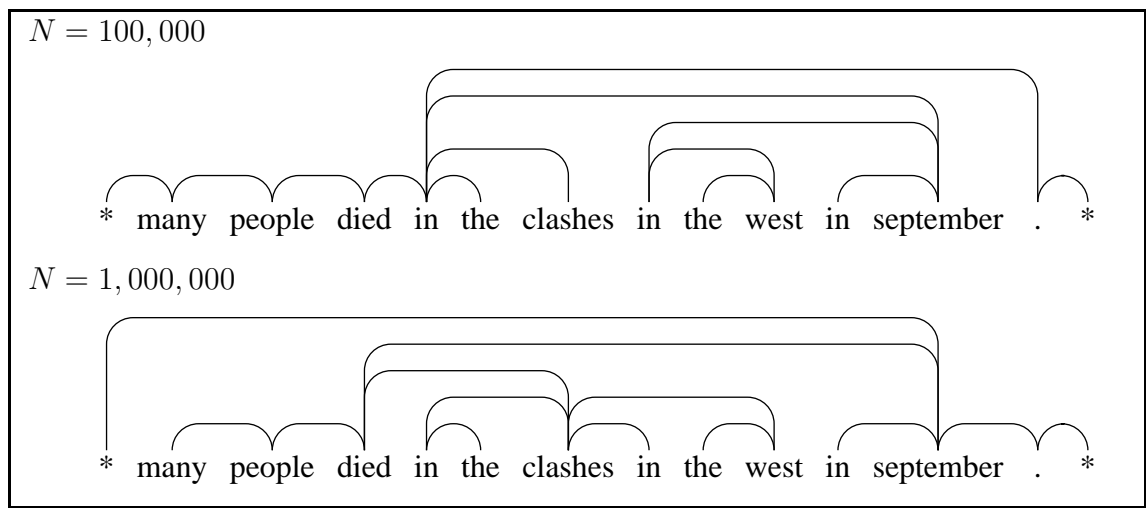

\centering
\fbox{
\parbox{5.8in}{
$N=100,000$

\myvspace{50pt}
\linkage
{[(*)(many)(people)(died)(in)(the)(clashes)(in)(the)(west)(in)(september)( . )( * )]}
{[[0 1 0 ()][1 2 0 ()][2 3 0 ()][3 4 0 ()][4 5 0 ()][4 6 1 ()][8 9 0 ()][7 9 1 ()][10 11 0 ()][7 11 2 ()][4 11 3 ()][4 12 4 ()][12 13 0 ()]]}
{[0 ]}

$N=1,000,000$

\myvspace{50pt}
\linkage
{[(*)(many)(people)(died)(in)(the)(clashes)(in)(the)(west)(in)(september)( . )( * )]}
{[[1 2 0 ()][2 3 0 ()][4 5 0 ()][4 6 1 ()][3 6 2 ()][6 7 0 ()][8 9 0 ()][6 9 1 ()][10 11 0 ()][3 11 3 ()][0 11 4 ()][11 12 0 ()][12 13 0 ()]]}
{[0 ]}
}}
\caption{Prepositional phrase attachment.}
\label{pp}
\end{figure}

Figure~\ref{pp} shows a prepositional phrase attachment
problem.  The sentence ends with three prepositional phrases, each
starting with the word ``in''.  Syntax does not uniquely determine
where they should be attached.  At 100,000 words, the program still
has not decided on the final attachment.  Somewhere between 100,000
words and 1,000,000 words, it learns enough to relate \word{died} to
\word{clashes}, \word{clashes} to \word{west}, and \word{september} to
\word{died}.  Note that {\em ``died in the west''} and {\em ``clashes
in september''} are also meaningful phrases.  However the links
discovered by the program had stronger attraction.

\begin{figure}[h]
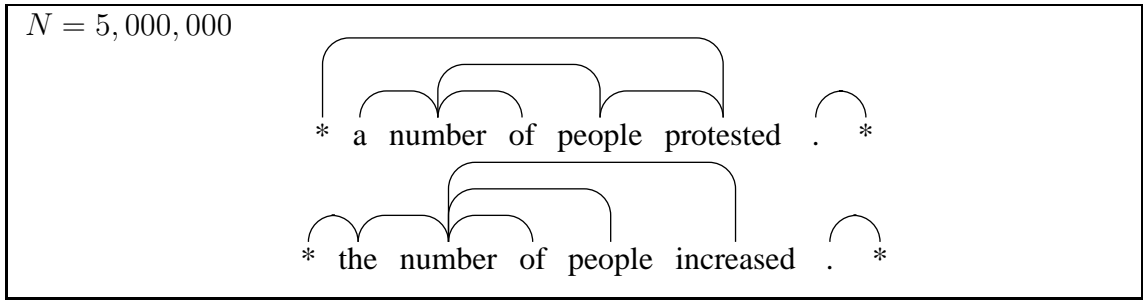

\centering
\fbox{
\parbox{5.8in}{
$N=5,000,000$

\myvspace{20pt}
\linkage
{[(*)(a)(number)(of)(people)(protested)( . )( * )]}
{[[1 2 0 ()][2 3 0 ()][2 4 1 ()][4 5 0 ()][0 5 2 ()][6 7 0 ()]]}
{[0 ]}

\myvspace{20pt}
\linkage
{[(*)(the)(number)(of)(people)(increased)( . )( * )]}
{[[0 1 0 ()][1 2 0 ()][2 3 0 ()][2 4 1 ()][2 5 2 ()][6 7 0 ()]]}
{[0 ]}
}}
\caption{Distinguishing syntactically identical sentences.}
\label{of}
\end{figure}

Figure~\ref{of} illustrates a common type of ambiguity
related to the of-phrase.  The English preposition \word{of} is
particularly ambiguous in its semantic function \cite{Quirk85}.  It
can be used in a function similar to that of the genitive ({\em the
gravity of the earth $\sim$ the earth's gravity}), or in partitive
constructions ({\em bottle of wine}) among others.

The two sentences in Figure~\ref{of} are syntactically
identical.  They both have the same phrase {\em ``number of people''}
as subject.  In the first one it is the people who are doing the
protesting, whereas in the second one, it is the number which is
increasing.  After five million words of training, the lexical
attraction information becomes sufficient to find the correct subject.

\begin{figure}[h]
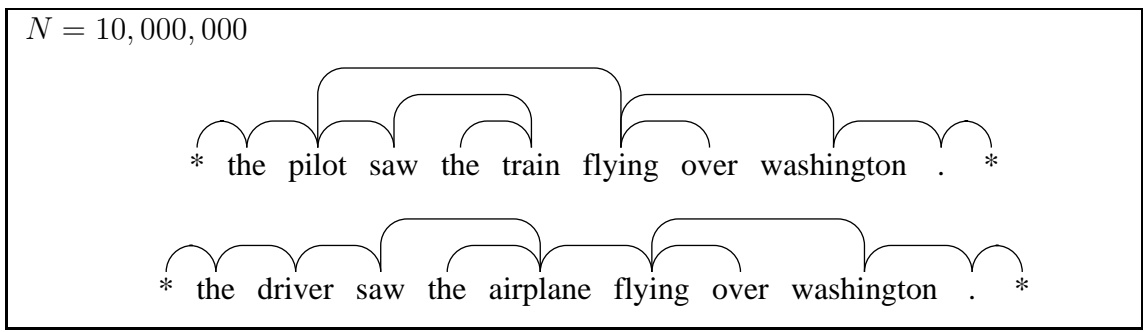

\centering
\fbox{
\parbox{5.8in}{
$N=10,000,000$

\myvspace{30pt}
\linkage
{[(*)(the)(pilot)(saw)(the)(train)(flying)(over)(washington)( . )( * )]}
{[[0 1 0 ()][1 2 0 ()][2 3 0 ()][4 5 0 ()][3 5 1 ()][2 6 2 ()][6 7 0 ()][6 8 1 ()][8 9 0 ()][9 10 0 ()]]}
{[0 ]}
\myvspace{20pt}
\linkage
{[(*)(the)(driver)(saw)(the)(airplane)(flying)(over)(washington)( . )( * )]}
{[[0 1 0 ()][1 2 0 ()][2 3 0 ()][4 5 0 ()][3 5 1 ()][5 6 0 ()][6 7 0 ()][6 8 1 ()][8 9 0 ()][9 10 0 ()]]}
{[0 ]}
}}
\caption{Who is flying?}
\label{fly}
\end{figure}

Figure~\ref{fly} presents our final example, which is analogous to
Sentence (\ref{sLenat}) from the previous chapter.  I replaced some
words with ones that were more frequent in the corpus.  The sentence
is ambiguous as to who is doing the flying.  The program is able to
link pilot with flying in the first case and airplane with flying in
the second case based on lexical attraction.

\chapter{Lexical Attraction Models}
\label{theory}

A probabilistic language model is a representation of linguistic
knowledge which assigns a probability distribution over the possible
strings of the language.  This chapter presents a new class of
language models called {\em lexical attraction models}.  Within the
framework of lexical attraction models it is possible to represent
syntactic relations which form the basis for extraction of meaning.
Syntactic relations are defined as pairwise relations between words
identifiable in the surface form of the sentence.  Lexical attraction
is formalized as the mutual information captured in a syntactic
relation.  The set of syntactic relations in a sentence is represented
as an undirected, acyclic, and planar graph\footnote{Syntactic
relation links drawn over words do not cross}.  I show that the
entropy of a lexical attraction model is determined by the mutual
information captured in the syntactic relations given by the model.
This is a prelude to the central idea of the next chapter: The search
for a low entropy language model leads to the unsupervised discovery
of syntactic relations.

\section{Syntactic relations are primitives of language}

A defining property of human language is compositionality, i.e. the
ability to construct compound meanings by combining simpler ones.  The
meaning of the sentence {\em ``John kissed Mary''} is more than just
a list of the three concepts \word{John}, \word{kiss}, and
\word{Mary}.  Part of the meaning is captured in the way these
concepts are brought together, as can be seen by combining them in a
different way: {\em ``Mary kissed John''} has a different meaning.

Language gives us the ability to describe concepts in certain semantic
relations by placing the corresponding words in certain syntactic
positions.  However, all languages are restricted to using a small
number of function words, affixes, and word order for the expression
of different relations because of the one dimensional nature of
language.  Therefore, the number of different syntactic arrangements
is limited.  I define the set of relations between words identifiable
in the syntactic domain as {\em syntactic relations}.  Examples of
syntactic relations are the subject-verb relation, the verb-object
relation and prepositional attachments.

Because of the limited number of possible syntactic relations there is
a many to one mapping from conceptual relations to syntactic
relations.  Compare the sentences {\em ``I saw the book''} and {\em
``I burnt the book''}.  The conceptual relation between \word{see} and
\word{book} is very different from the one between \word{burn} and
\word{book}.  For the verb \word{see}, the action has no effect on the
object, whereas for \word{burn} it does.  Nevertheless they have to be
expressed using the same verb-object relation.  The different types of
syntactic relations constrain what can be expressed and distinguished
in language.  Therefore, I take syntactic relations as the
representational primitives of language.

A phrase-structure representation of a sentence reveals which words or
phrases combine to form higher order units.  The actual syntactic
relations between words are only implicit in these groupings.  In a
phrase-structure formalism, the fact that \word{John} is the subject
of \word{kissed} can only be expressed by saying something like ``John
is the head of a noun-phrase which is a direct constituent of a
sentence which has a verb-phrase headed by the verb kissed.''

I chose to use syntactic relations as representational primitives for
two reasons.  First, the indirect representation of phrase-structure
makes unsupervised language acquisition very difficult.  Second, if
the eventual goal is to extract meaning, then syntactic relations are
what we need, and phrase-structure only indirectly helps us retrieve
them.  Instead of assuming that syntactic relations are an indirect
by-product of phrase-structure, I chose to take syntactic relations as
the basic primitives, and treat phrase-structure as an epiphenomenon
caused by trying to express syntactic relations within the one
dimensional nature of language.

The linguistic formalism that takes syntactic relations between words
as basic primitives is known as the dependency formalism.  Mel'\v{c}uk
discusses important properties of syntactic relations in his book on
dependency formalism \cite{Melcuk88}.  A large scale implementation of
English syntax based on a similar formalism by Sleator and Temperley
uses 107 different types of syntactic relations such as subject-verb,
verb-object, and determiner-noun \cite{Sleator91}.

I did not differentiate between different types of syntactic
relations in this work.  The goal of the learning program described in
Chapter~\ref{learning} is to correctly determine whether or not two
words in a sentence are syntactically related.  Learning to
differentiate between different types of syntactic relations in an
unsupervised manner is discussed in that chapter.

\section{Lexical attraction is the likelihood of a syntactic relation}

In order to understand a sentence, one needs to identify which words
in the sentence are syntactically related.  {\em Lexical attraction}
is the likelihood of a syntactic relation.  In this section I
formalize lexical attraction within the framework of information
theory.

Shannon defines the entropy of a discrete random variable as $H =
-\sum p_i \log p_i$ where $i$ ranges over the possible values of the
random variable and $p_i$ is the probability of value $i$
\cite{Shannon48,Cover91}.  Consider a sequence of tokens drawn
independently from a discrete distribution.  In order to construct the
shortest description of this sequence, each token $i$ must be encoded
using $-\log_2 p_i$ bits on average.  $-\log_2 p_i$ can be defined as
the information content of token $i$.  Entropy can then be interpreted
as the average information per token.  Following is an English
sentence with the information content of each word\footnote{The word
probabilities were estimated using a 227 million word corpus dominated
by news material.} given below, assuming words are independently
selected.  Note that the information content is lower for the more
frequently occurring words.

\begin{sentence}
\label{ira}
\mylinkage{[[(The)(4.20)][(IRA)(15.85)][(is)(7.33)][(fighting)(13.27)][(British)(12.38)][(rule)(13.20)][(in)(5.80)][(Northern)(12.60)][(Ireland)(14.65)]]}{[]}
\end{sentence}

Why do we care about encoding or compression?  The total information
in sentence (\ref{ira}) is the sum of the information of each word,
which is 99.28 bits.  This is mathematically equivalent to the
statement that the probability of seeing this sentence is the product
of the probabilities of seeing each word, which is $2^{-99.28}$.
Therefore there is an equivalence between the entropy and the
probability assigned to the input.  A probabilistic language model
assigns a probability distribution over all possible sentences of the
language.  The maximum likelihood principle states that the parameters
of a model should be estimated so as to maximize the probability
assigned to the observed data.  This means that the most likely
language model is also the one that achieves the lowest entropy.

A model can achieve lower entropy only by taking into account the
relations between the words in the sentence.  Consider the phrase {\em
Northern Ireland}.  Even though the independent probability of
\word{Northern} is $2^{-12.6}$, it is seen before \word{Ireland} 36\%
of the time.  Another way of saying this is that although
\word{Northern} carries 12.6 bits of information by itself, it adds
only 1.48 bits of new information to \word{Ireland}.

With this dependency, \word{Northern} and \word{Ireland} can be
encoded using $1.48+14.65=16.13$ bits instead of $12.60+14.65=27.25$
bits.  The 11.12 bit gain from the correlation of these two words is
called mutual information.  Lexical attraction is measured with mutual
information.  The basic assumption of this work is that words with
high lexical attraction are likely to be syntactically related.

\section{The context of a word is given by its syntactic relations}

The {\em Northern Ireland} example shows that the information content
of a word depends on other related words, i.e. its context.  The
context of a word in turn is determined by the language model used.
In this section, I describe a new class of language models called {\em
lexical attraction models}.  Within the lexical attraction framework,
it is possible to represent a linguistically plausible context for a
word.

The choice of context by a language model implies certain
probabilistic independence assumptions.  For example, an n-gram model
defines the context of a word as the $n-1$ words immediately preceding
it.  Diagram (\ref{ira2}) gives the information content of the words
in (\ref{ira}) according to a bigram model.  The arrows show the
dependencies.  The information content of {\em Northern} and {\em
Ireland} is different from the previous section because of the
different dependencies.  The assumption is that each word is
conditionally independent of everything but the adjacent words.  The
information content of each word is computed based on its conditional
probability given the previous word.  As a result, the encoding of the
sentence is reduced from 99.28 bits to 62.34 bits.

\vspace{10pt}
\begin{sentence}
\label{ira2}
\mylinkage
{[[(The)(4.20)][(IRA)(12.90)][(is)(3.73)][(fighting)(10.54)][(British)(8.66)][(rule)(5.96)][(in)(3.57)][(Northern)(9.25)][(Ireland)(3.53)]]}
{[[0 1 0 (>)][1 2 0 (>)][2 3 0 (>)][3 4 0 (>)][4 5 0 (>)][5 6 0 (>)][6 7 0 (>)][7 8 0 (>)]]}
\end{sentence}

Two words in a sentence are almost never completely independent.  In
fact, Beeferman et al. report that words can continue to show
selectional influence for a window of several hundred words
\cite{Beeferman97a}.  However, the degree of the dependency falls
exponentially with distance.  That justifies the choice of the n-gram
models to relate dependency to proximity.

Nevertheless, using the previous $n-1$ words as context is against our
linguistic intuition.  In a sentence like {\em ``The man with the dog
spoke''}, the selection of \word{spoke} is influenced by \word{man}
and is independent of the previous word \word{dog}.  It follows that
the context of a word would be better determined by its linguistic
relations rather than according to a fixed pattern.

Words in direct syntactic relation have strong dependencies.  Chomsky
defines such dependencies as {\em selectional relations}
\cite{Chomsky65}.  Subject and verb, for example, have a selectional
relation, and so do verb and object.  Subject and object, on the other
hand, are assumed to be chosen independently of one another.  It
should be noted that this independence is only an approximation.  The
sentences {\em ``The doctor examined the patient''} and {\em ``The
lawyer examined the witness''} show that the subject can have a strong
influence on the choice of the object.  Such second degree effects
are discussed in Chapter~\ref{learning}.

The following diagram gives the information content of the words in
sentence (\ref{ira}) based on direct syntactic relations:

\vspace{20pt}
\begin{sentence}
\label{ira3}
\mylinkage
{[[(The)(1.25)][(IRA)(6.60)][(is)(4.60)][(fighting)(13.27)][(British)(5.13)][(rule)(8.13)][(in)(2.69)][(Northern)(1.48)][(Ireland)(6.70)]]}
{[[0 1 0 (<)][1 3 1 (<)][2 3 0 (<)][3 5 1 (>)][4 5 0 (<)][5 8 2 (>)][6 8 1 (<)][7 8 0 (<)]]}
\end{sentence}

The arrows represent the head-modifier relations between words.  The
information content of each word is computed based on its conditional
probability given its head.  I marked the verb as governing the
auxiliary and the noun governing the preposition which may look
controversial to linguists.  From an information theory perspective,
the mutual information between content words is higher than that of
function words.  Therefore my model does not favor function word
heads.

The probabilities were estimated by counting the occurrences of each
pair in the same relative position.  The linguistic dependencies
reduce the encoding of the words in this sentence to 49.85 bits
compared to the 62.34 bits of the bigram model\footnote{These numbers
do not take into account the encoding of the dependency structure.}.

Every model has to make certain independence assumptions, otherwise
the number of parameters would be prohibitively large to learn.  The
choice of the independence assumptions determine the context of a
word.  The assumption in (\ref{ira3}) is that each word depends on one
other word in the sentence, but not necessarily an adjacent word as in
n-gram models.  I define the class of language models that are based
on this assumption as {\em lexical attraction models}.  Lexical
attraction models make it possible to define the context of the word
in terms of its syntactic relations.

\section{Entropy is determined by syntactic relations}

I define the set of probabilistic dependencies imposed by syntactic
relations in a sentence as the {\em dependency structure} of the
sentence.  The strength of the links in a dependency structure is
determined by lexical attraction.  In this section I formalize
dependency structures as Markov networks and show that the entropy of
a language model is determined by the mutual information captured in
syntactic relations.

A Markov network is an undirected graph representing the joint
probability distribution for a set of variables\footnote{As opposed to
Bayesian networks which are directed.} \cite{Pearl88}.  Each vertex
corresponds to a random variable, a word in our case.  The structure
of the graph represents a set of conditional independence properties
of the distribution: each variable is probabilistically independent of
its non-neighbors in the graph given the state of its neighbors.

You can see two interesting properties of the dependency structure in
diagram~(\ref{ira3}): The graph formed by the syntactic relations is
acyclic and the links connecting the words do not cross.  In this
section you will also see that the directions of the links do not
effect the joint probability of the sentence, thus the links can be
undirected.  I discuss these three properties below and derive a
formula for the entropy of the model.

\subsubsection*{Dependency structure is acyclic}

The syntactic relations in a sentence form a tree.  Trees are acyclic.
Linguistically, each word in a sentence has a unique governor, except
for the head word, which governs the whole sentence\footnote{See
\cite[p.\ 25]{Melcuk88} for a discussion.}.  If you assume
that each word probabilistically depends on its governor, the
resulting dependency structure will be a rooted tree as in
diagram~(\ref{ira3}).

\subsubsection*{Dependency structure is planar}

Most sentences in natural languages have the property that syntactic
relation links drawn over words do not cross.  This property is called
{\em planarity} \cite{Sleator93}, {\em projectivity} \cite{Melcuk88},
or {\em adjacency} \cite{Hudson84} by various researchers.  The
examples below illustrate the planarity of English.  In
sentence~(\ref{planar1}), it is easily seen that the woman was in the
red dress and the meeting was in the afternoon.  However, in
sentence~(\ref{planar2}) the same interpretation is not possible.  In
fact, it seems more plausible for John to be in the red dress.

\vspace{10pt}
\begin{sentence} \label{planar1}
\mylinkage{[[(John)()][(met)()][(the)()][(woman)()][(in)()][(the)()][(red)()][(dress)()][(in)()][(the)()][(afternoon)()]]}{[[3 7 0 ()][1 10 1 ()]]}
\end{sentence}

\begin{sentence} \label{planar2}
\mylinkage{[[(John)()][(met)()][(the)()][(woman)()][(in)()][(the)()][(afternoon)()][(in)()][(the)()][(red)()][(dress)()]]}{[[1 6 0 ()][3 10 1 (?)]]}
\end{sentence}

Gaifman gave the first formal analysis of dependency structures that
satisfy the planarity condition\cite{Gaifman65}.  His paper gives a
natural correspondence between dependency systems and phrase-structure
systems and shows that the dependency model characterized by planarity
is context-free.  Sleator and Temperley show that their planar model
is also context-free even though it allows cycles \cite{Sleator91}.

\subsubsection*{Lexical attraction is symmetric}

Lexical attraction between two words is symmetric.  The mutual
information is the same no matter which direction the dependency goes.
This directly follows from Bayes' rule.  What is less obvious is that
the choice of the head word and the corresponding dependency
directions it imposes do not effect the joint probability of the
sentence.  The joint probability is determined only by the choice of
the pairs of words to be linked.  Therefore dependency structures can
be formalized as Markov networks, i.e. they are undirected.

Consider the {\em Northern Ireland} example:

\begin{sentence}
\mylinkage{[[(Northern)(1.48)][(Ireland)(14.65)]]}{[[0 1 0 (<)]]}
\hspace{3cm}
\mylinkage{[[(Northern)(12.60)][(Ireland)(3.53)]]}{[[0 1 0 (>)]]}
\end{sentence}

In the first case, I used the conditional probability of {\em
Northern} given that the next word is {\em Ireland}.  In the second
case, I used the conditional probability of {\em Ireland} given that
the previous word is {\em Northern}.  In both cases the encoding of
the two words is 16.13 bits, which is in fact $-\log_2 p$ of the joint
probability of {\em Northern Ireland}.  Thus a more natural
representation would be (\ref{newlink}), where the link has no
direction and its label shows the number of bits gained, mutual
information:

\vspace{10pt}
\begin{sentence}
\label{newlink}
\mylinkage{[[(Northern)(12.60)][(Ireland)(14.65)]]}{[[0 1 0 (11.12)]]}
\end{sentence}

I generalize this result below and use the same representation for the
whole sentence:

\vspace{30pt}
\begin{sentence}
\label{ira4}
\mylinkage
{[[(The)(4.20)][(IRA)(15.85)][(is)(7.33)][(fighting)(13.27)][(British)(12.38)][(rule)(13.20)][(in)(5.80)][(Northern)(12.60)][(Ireland)(14.65)]]}
{[[0 1 0 (2.95)][1 3 1 (9.25)][2 3 0 (2.73)][3 5 1 (5.07)][4 5 0 (7.25)][5 8 2 (7.95)][6 8 1 (3.11)][7 8 0 (11.12)]]}
\end{sentence}

\begin{theorem}
The probability of a sentence with a given dependency structure does
not depend on the choice of the head word.
\end{theorem}

\begin{proof}
Consider a sentence $S$ where:
\begin{eqnarray*}
W &=& \{w_0, w_1, \ldots, w_n\} \\
L &=& \{(w_{i1},w_{j1}), (w_{i2},w_{j2}), \ldots\}
\end{eqnarray*}
denote words and links respectively.  Let $P(L)$ denote the
probability of a sentence having the dependency structure given by
$L$.  Assume that $w_0$ is the head word and every word
probabilistically depends only on its governor.  Then the joint
probability of the sentence is given by the following expression:

\begin{sentence}
\label{undirected}
\parbox{3in}{
\begin{eqnarray*}
P(S) & = & P(L) P(w_0) \prod_{(w_i,w_j)\in L} P(w_j \mid w_i) \\
& = & P(L) P(w_0) \prod_{(w_i,w_j)\in L} \frac{P(w_i,w_j)}{P(w_i)} \\
& = & P(L) \prod_{w_i\in W} P(w_i)
\prod_{(w_i,w_j)\in L} \frac{P(w_i,w_j)}{P(w_i) P(w_j)}
\end{eqnarray*}
}
\end{sentence}

In the final expression $P(w_0)$ plays no special role, i.e. starting
from any other head word, I would have arrived at the same result.
Therefore the choice of the head and the corresponding directions
imposed on the links are immaterial for the probability of the
sentence.
\end{proof}

\subsubsection*{Encoding of dependency structure is linear}

The $P(L)$ factor in (\ref{undirected}) represents the prior
probability that the language model assigns to a dependency structure.
In n-gram models $P(L)$ is always 1, because there is only one
possible dependency structure.  In probabilistic context free
grammars, the probabilities assigned to grammar rules impose a
probability distribution over possible parse trees.  In my model, I
assume a uniform distribution over all possible dependency structures,
i.e.  $P(L) = 1/|L|$, where $|L|$ is the number of possible dependency
structures.

Without the planarity condition, the number of possible dependency
structures for an n word sentence would be given by Cayley's formula:
$n^{n-2}$ \cite{Harary69}.  The encoding of the dependency structure
would then take $O(n\log n)$ bits.  However, the encoding of planar
dependency structures is linear in the number of words as the
following theorem shows.

\begin{theorem} \label{ntree}
Let $f(n)$ be the number of possible dependency structures for an
$n+1$ word sentence.  We have:
\[f(n) = \frac{1}{2n+1} \choose{3n}{n}\]
\end{theorem}

\begin{proof}
Consider a sentence with $n+1$ words as in (\ref{recurrence}).  The
leftmost word $w_0$ must be connected to the rest of the sentence via
one or more links.  Even though the links are undirected, I will
impose a direction taking $w_0$ as the head to make the argument
simpler.  Let $w_i$ be the leftmost child of $w_0$.  We can split the
rest of the sentence into three groups: Left descendants of $w_i$ span
$w_1^{i-1}$, right descendants of $w_i$ span $w_{i+1}^j$, and the rest
of the sentence spans $w_{j+1}^n$.  Each of these three groups can be
empty.  The notation $w_i^j$ denotes the span of words $w_i \ldots
w_j$.

\begin{sentence}
\label{recurrence}
\begin{picture}(350,50)
\put(0,0){$w_0$}
\put(40,0){$w_1$}
\put(60,0){$\ldots$}
\put(80,0){$w_{i-1}$}
\put(130,0){$w_{i}$}
\put(170,0){$w_{i+1}$}
\put(200,0){$\ldots$}
\put(220,0){$w_j$}
\put(260,0){$w_{j+1}$}
\put(290,0){$\ldots$}
\put(310,0){$w_n$}
\put(70,0){\oval(80,20)}
\put(200,0){\oval(80,20)}
\put(290,0){\oval(80,20)}
\put(100,10){\oval(70,20)[t]}
\put(170,10){\oval(70,20)[t]}
\put(70,10){\oval(130,40)[t]}
\put(150,10){\oval(290,60)[t]}
\end{picture}

\end{sentence}

The problem of counting the number of dependency structures for the
$n$ words headed by $w_0$ can be split into three smaller versions of
the same problem: count the number of structures for $w_1^{i-1}$
headed by $w_i$, $w_{i+1}^j$ headed by $w_i$, and $w_{j+1}^n$ headed
by $w_0$.  Therefore $f(n)$ can be decomposed with the following
recurrence relation:

\[ f(n) = \sum_{p+q+r=n-1} f(p) f(q) f(r)
\mbox{\hspace{1cm}} p,q,r \ge 0 \]

Here the numbers $p$, $q$, and $r$ represent the number of words in
$w_1^{i-1}$, $w_{i+1}^j$, and $w_{j+1}^n$ respectively.  This is a
recurrence with 3-fold convolution.  The general expression for a
recurrence with m-fold convolution is $C(mn,n)/(mn-n+1)$ where $C$ is
the binomial coefficient \cite[p.\ 361]{Graham94}.  Therefore $f(n) =
C(3n,n)/(2n+1)$.
\end{proof}

\begin{figure}[ht]
\centering
\fbox{\psfig{file=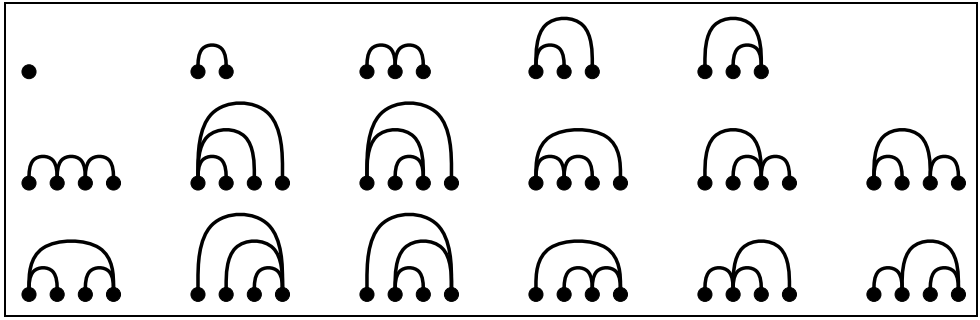,width=5in}}
\caption{Possible dependency structures with up to four words}
\label{dtree}
\end{figure}

The first few values of $f(n)$ are: 1, 1, 3, 12, 55, 273, 1428.
Figure~\ref{dtree} shows the possible dependency structures with up to
four words.

An upper bound on the number of dependency structures can be obtained
using the following inequality given in \cite[p.\ 102]{Cormen90}:

\begin{eqnarray*}
\choose{n}{k} & \le & \frac{n^n}{k^k (n-k)^{n-k}} \\
\choose{3n}{n} & \le & \frac{(3n)^{3n}}{(n)^{n} (2n)^{2n}} \\
& = & \frac{3^{3n}}{2^{2n}}
\end{eqnarray*}

Taking the logarithm of this value and dividing it by $n$, you can see
that the encoding of a planar dependency structure takes less than
$3\log_2 3 - 2 \approx 2.75$ bits per word.

\subsubsection*{Entropy is determined by syntactic relations}

With these results at hand, it is revealing to look at
(\ref{undirected}) from an information theory perspective.  The final
expression of (\ref{undirected}) can be rewritten as:

\[ -\log_2 P(S) = - \sum_{w_i\in W} \log_2 P(w_i) - \log_2 P(L) 
- \sum_{(w_i,w_j)\in L} \frac{P(w_i,w_j)}{P(w_i) P(w_j)} \]

\noindent This can be interpreted as:

\begin{sentence}
\label{entropy} \samepage \em 
information in a sentence = information in the words \\
\hspace*{2cm} + information in the dependency structure \\
\hspace*{2cm} -- mutual information captured in syntactic relations.
\end{sentence}

The average information of an isolated word is independent of the
language model.  I also showed that the encoding of the dependency
structure is linear in the number of words for lexical attraction
models.  Therefore the first two terms in (\ref{entropy}) have a
constant contribution per word and the entropy of the model is
completely determined by the mutual information captured in syntactic
relations.

\section{Summary}

Part of the meaning in language is captured in the way words are
arranged in a sentence.  This arrangement implies certain syntactic
relations between words.  With a view towards extraction of meaning, I
adopted a linguistic representation that takes syntactic relations as
its basic primitives.

I defined lexical attraction as the likelihood of two words being
related.  The language model determines which words are related in a
sentence.  For example, n-gram models assume that each word depends on
the previous n-1 words.  I defined a new class of language models
called lexical attraction models where each word depends on one other
word in the sentence which is not necessarily adjacent, provided that
link crossing and cycles are not allowed.  It is possible to represent
linguistically plausible head modifier relationships within this
framework.  I showed that the entropy of such a model is determined by
the lexical attraction between related words.

The examples in this chapter showed that using linguistic relations
between words can lead to lower entropy.  Conversely, the search for a
low entropy language model can lead to the unsupervised discovery of
linguistic relations, as I will show in the next chapter.

\chapter{Bootstrapping Acquisition}
\label{learning}

\begin{figure}[h]
\centering
\fbox{\psfig{file=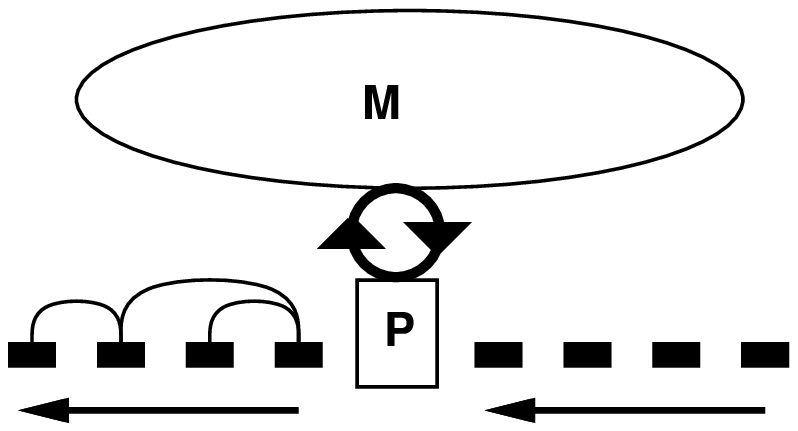,height=2in}}
\caption{Interdigitating learning and processing to bootstrap acquisition.}
\label{bootstrap2}
\end{figure}

This chapter presents a bootstrapping mechanism for learning to
identify linguistic relations.  The key to the mechanism is to
interdigitate learning and processing.  That way the structures built
by the processor can help the learner detect higher level
regularities.  Figure~\ref{bootstrap2} depicts this feedback loop
between the memory and the processor.

In the beginning of the training process, the processor cannot build
any structure and the memory is presented with the raw input.  As the
memory detects low level correlations in the raw input, the processor
uses this information to assign simple structure.  The simple
structure enables the memory to detect higher level correlations,
which in turn enables the processor to assign more complex structure.
Using this bootstrapping mechanism, the program starts reading raw
text with no initial knowledge and reaches 60\% precision and 50\%
recall in content word links after training.

I describe the implementation of the processor in the first section
and the memory in the second section.  This is followed by
quantitative results of the learning process and a critical evaluation
of its shortcomings.  The final section discusses related work.

The key insights that differentiate my approach from others are:

\begin{itemize}
\item Training with words instead of parts of speech enables the
program to learn common but idiosyncratic usages of words.
\item Not committing to early generalizations prevent the program from
making irrecoverable mistakes.
\item Using a representation that makes the relevant features explicit
simplifies learning.  I believe what makes humans good learners is not
sophisticated learning algorithms but having the right representations.
\end{itemize}

\section{Processor}

The goal of the processor is to find the dependency structure that
assigns a given sentence a high probability.  In Chapter~\ref{theory},
I showed that the probability of a sentence is determined by the
mutual information captured in syntactic relations.  Thus, the problem
is to find the dependency structure with the highest total mutual
information.  The optimum solution can be found in time $O(n^5)$.  I
use an approximation algorithm that is $O(n^2)$.  The main reason for
this choice was the simplicity of the resulting processor as well as
its speed.  The simplicity is important because in the architecture of
Figure~\ref{bootstrap2}, the input to the memory consists of the
states and the actions of the processor, rather than the raw input
signal.  In order to make learning easy, the processor should be
simple, i.e. it should have a small number of possible states and a
small number of possible actions.  Below, I present both the
approximation algorithm and the optimum algorithm.

\subsubsection*{Approximation algorithm}

There are two possible actions of the simple processor: given two
words, they can be linked or not linked.  The two words under
consideration constitute the state of the processor visible to the
memory.  The order in which words come under the processor's attention
is determined by a simple control system.

The control system reads the words from left to right.  After reading
each new word, the processor tries to link it to each of the previous
words.  When a link crossing or a cycle is detected, the weakest links
in conflict is eliminated.  Following is the pseudo-code for this
$n^2$ algorithm:

\bigskip
\codebox{
\<$\proc{Link-Sentence}(S)$ \\
\li \For $j \gets 1$ \To $\proc{Length}(S)$ \\
\li \> \For $i \gets j-1$ \Downto $1$ \\
\li \> \> $\id{last} \gets \proc{Pop}(\proc{Right-Links}(i),{\sc stack})$ \\
\li \> \> ${\sc minlink}[i] \gets 
\proc{Min}(\id{last}, {\sc minlink}[\proc{Right-Word}(\id{last})]) $ \\
\li \> \> \If $\proc{MI}(S[i],S[j]) > 0$ \\
\li \> \> \> {\bf and} $\proc{MI}(S[i],S[j]) > 
\proc{MI}({\sc minlink}[i])$ \\
\li \> \> \> {\bf and} $\forall s:\proc{MI}(S[i],S[j]) >
\proc{MI}({\sc stack}[s])$ \\
\li \> \> \Then $\forall s: \proc{Unlink}({\sc stack}[s])$ \\
\li \> \> \> $\proc{Reset}({\sc stack})$ \\
\li \> \> \> $\proc{Unlink}({\sc minlink}[i])$ \\
\li \> \> \> ${\sc minlink}[i] \gets \proc{Link}(S[i],S[j])$ \\
\li \> \> $\proc{Push}(\proc{Left-Links}(i),{\sc stack})$ \\
}

The input $S$ to the procedure \proc{Link-Sentence} is the sentence as
an array of words.  The links are created and deleted using
\proc{Link} and \proc{Unlink}.  The function \proc{MI} gets the mutual
information value for a link from the memory.  The {\sc minlink} array
and the {\sc stack} are used to detect cycles and link crossings.  The
$i$ th element of {\sc minlink} contains the minimum valued link on
the path from $S[i]$ to $S[j]$ if there exists one.  As $i$ moves
leftward in the sentence, the right links of $S[i]$ are popped and the
left links of $S[i]$ are pushed to the {\sc stack}.  The last right
link popped or the {\sc minlink} on its right side becomes the new
${\sc minlink}[i]$.  A link crossing is detected when a new link is
created with a nonempty stack.  A cycle is detected when a new link is
created when ${\sc minlink}[i]$ is nonempty.

Note that when a strong new link crosses multiple weak links, it is
accepted and the weak links are deleted even if the new link is weaker
than the sum of the old links.  Although this action results in lower
total mutual information, it was implemented because multiple weak
links connected to the beginning of the sentence often prevented a
strong meaningful link from being created.  This way the directional
bias of the approximation algorithm was partly compensated for.

To get a sense of how the link crossing and cycle constraints,
combined with the knowledge of lexical attraction, lead to the
identification of linguistic relations, a sequence of snapshots of the
processor running on a simple sentence is presented below.  Note that
one or more steps may be skipped between two snapshots:

\begin{itemize}

\item Words are read from left to right:

\vspace*{20pt}\hspace{1cm}\mylinkage
{[[(*)()][(these)()]]}
{[[0 1 0 ([1.18])]]}

\item Each word checks the words to its left for possible links.  The
link under consideration is marked with its value in square brackets.
The accepted links have no square brackets:

\vspace*{20pt}\hspace{1cm}\mylinkage
{[[(*)()][(these)()][(people)()]]}
{[[0 1 0 (1.18)][1 2 0 ([3.48])]]}

\item The algorithm detects cycles and eliminates the weakest link in
the cycle:

\vspace*{30pt}\hspace{1cm}\mylinkage
{[[(*)()][(these)()][(people)()]]}
{[[0 1 0 (1.18)][1 2 0 (3.48)][0 2 1 ([0.55])]]}

\item Negative links are not accepted:

\vspace*{20pt}\hspace{1cm}\mylinkage
{[[(*)()][(these)()][(people)()][(also)()]]}
{[[0 1 0 (1.18)][1 2 0 (3.48)][2 3 0 ([-1.64])]]}

\item The algorithm detects crossing links:

\vspace*{40pt}\hspace{1cm}\mylinkage
{[[(*)()][(these)()][(people)()][(also)()][(want)()]]}
{[[0 1 0 (1.18)][1 2 0 (3.48)][3 4 0 (1.43)][0 3 1 (1.78)][2 4 2 ([3.15])]]}

\item The weaker conflicting link gets eliminated:

\vspace*{20pt}\hspace{1cm}\mylinkage
{[[(*)()][(these)()][(people)()][(also)()][(want)()]]}
{[[0 1 0 (1.18)][1 2 0 (3.48)][2 4 1 (3.15)][3 4 0 (1.43)]]}

\item The combination of the link crossing and cycle constraints and
the knowledge of lexical attraction help eliminate links that do not
correspond to syntactic relations.  In this diagram \word{more}
strongly attracts \word{money}, which will result in the elimination
of the meaningless link between * and \word{government}:

\vspace*{50pt}\hspace{1cm}\mylinkage
{[[(*)()][(these)()][(people)()][(also)()][(want)()][(more)()][(government)()][(money)()]]}
{[[0 1 0 (1.18)][1 2 0 (3.48)][2 4 1 (3.15)][2 5 2 (1.26)][3 4 0 (1.43)][0 6 3 (0.53)][6 7 0 (0.43)][5 7 4 ([4.01])]]}

\item \word{Want} and \word{money} are also strongly attracted and
their link replaces the one between \word{people} and \word{more}:

\vspace*{40pt}\hspace{1cm}\mylinkage
{[[(*)()][(these)()][(people)()][(also)()][(want)()][(more)()][(government)()][(money)()]]}
{[[0 1 0 (1.18)][1 2 0 (3.48)][2 4 1 (3.15)][2 5 2 (1.26)][3 4 0 (1.43)][6 7 0 (0.43)][5 7 1 (4.01)][4 7 3 ([2.09])]]}

\item The bad links have been eliminated.

\vspace*{30pt}\hspace{1cm}\mylinkage
{[[(*)()][(these)()][(people)()][(also)()][(want)()][(more)()][(government)()][(money)()]]}
{[[0 1 0 (1.18)][1 2 0 (3.48)][2 4 1 (3.15)][3 4 0 (1.43)][6 7 0 (0.43)][5 7 1 (4.01)][4 7 2 (2.09)]]}

\item In the first cycle example, the new link was rejected because it
was weak.  In this example the new link is strong and it eliminates
one of the old links in the cycle.

\vspace*{30pt}\hspace{1cm}\mylinkage
{[[(*)()][(these)()][(people)()][(also)()][(want)()][(more)()][(government)()][(money)()][(for)()][(education)()]]}
{[[0 1 0 (1.18)][1 2 0 (3.48)][2 4 1 (3.15)][3 4 0 (1.43)][6 7 0 (0.43)][5 7 1 (4.01)][4 7 2 (2.09)][7 8 0 (2.61)][8 9 0 (2.58)][7 9 1 ([3.92])]]}

\item This is the final result:

\vspace*{30pt}\mylinkage
{[[(*)()][(these)()][(people)()][(also)()][(want)()][(more)()][(government)()][(money)()][(for)()][(education)()][(.)()][(*)()]]}
{[[0 1 0 (1.18)][1 2 0 (3.48)][2 4 1 (3.15)][3 4 0 (1.43)][6 7 0 (0.43)][5 7 1 (4.01)][4 7 2 (2.09)][7 8 0 (2.61)][7 9 1 (3.92)][9 10 0 (1.07)][10 11 0 (4.51)]]}

\end{itemize}

This algorithm is not guaranteed to find the most likely linkage.
Also it can leave some words disconnected.  Section~\ref{critical}
argues that the limiting factor for the performance of the program is
the accuracy of the learning process due to representational
limitations.  Thus the marginal gain from an optimal algorithm would
not be significant.  The algorithm presented in this section performs
reasonably well on the average and its speed and simplicity make it a
good candidate for training.

\subsubsection*{Optimal algorithm}

A Viterbi style algorithm \cite{Viterbi67} that finds the dependency
structure with the highest mutual information can be designed based on
the decomposition given in the proof of Theorem~\ref{ntree}.  The
relevant figure is reproduced below for convenience:

\bigskip
\begin{picture}(350,50)
\put(0,0){$w_0$}
\put(40,0){$w_1$}
\put(60,0){$\ldots$}
\put(80,0){$w_{i-1}$}
\put(130,0){$w_{i}$}
\put(170,0){$w_{i+1}$}
\put(200,0){$\ldots$}
\put(220,0){$w_j$}
\put(260,0){$w_{j+1}$}
\put(290,0){$\ldots$}
\put(310,0){$w_n$}
\put(70,0){\oval(80,20)}
\put(200,0){\oval(80,20)}
\put(290,0){\oval(80,20)}
\put(100,10){\oval(70,20)[t]}
\put(170,10){\oval(70,20)[t]}
\put(70,10){\oval(130,40)[t]}
\put(150,10){\oval(290,60)[t]}
\end{picture}

Let $\sigma(h,a,b)$ be the dependency structure that gives the highest
mutual information for the span $w_a^b$ dominated by a word $w_h$
outside this span.  For convenience, I assume that the leftmost word
dominates the whole sentence as in Theorem~\ref{ntree}.  Thus, the
optimal algorithm must find $\sigma(0,1,n)$.  Let $\alpha(w_a^b |
w_h)$ be the probability $\sigma(h,a,b)$ assigns to the span $w_a^b$
dominated by $w_h$.  For spans of length 1 and 0, $\alpha$ is given
by:

\begin{eqnarray*}
\alpha(w_a^a | w_h) & = & P(w_a | w_h) \makebox[20pt]{} h \ne a \\
\alpha(w_a^b | w_h) & = & 1 \makebox[60pt]{} b < a
\end{eqnarray*}

Given the $\alpha$ values for spans of length up to $l-1$, the
algorithm can compute it for spans of length $l$ as follows:

\begin{eqnarray*}
\alpha(w_a^b | w_h) 
& = & \max_{i,j} P(w_i | w_h) 
\alpha(w_a^{i-1} | w_i)
\alpha(w_{i+1}^j | w_i)
\alpha(w_{j+1}^b | w_h)
\makebox[20pt]{} h < a \le i \le j \le b \\
& = & \max_{i,j} P(w_i | w_h) 
\alpha(w_{i+1}^b | w_i)
\alpha(w_j^{i-1} | w_i)
\alpha(w_a^{j-1} | w_h)
\makebox[20pt]{} a \le j \le i \le b < h \\
\end{eqnarray*}

Thus, the algorithm can compute the $\alpha$ values bottom up,
starting with shorter spans and computing the longer spans using the
previous $\alpha$ values.  For each $\alpha$ value, the corresponding
$\sigma$ structure can be recorded.  At the end the structure
$\sigma(0,1,n)$ gives the answer.

The recursive computation for $\alpha(w_a^b | w_h)$ takes $O(n^2)$
steps.  For each length $l$ from 1 to $n$, $\alpha$ must be computed
for every span of length $l$ and every possible head for each span,
which means $O(n^3)$ $\alpha$ computations.  Thus the total
computation is $O(n^5)$.

\section{Memory}

The memory is a store of lexical attraction information.  The lexical
attraction of a word pair is computed based on the frequency with
which the pair comes to the processor's attention.  This information
is then used by the processor when deciding whether to link two
words.  In this section I describe three different procedures for
updating the memory.  In the next section, I present the performance
results of these three procedures.

The memory may record pairs when the processor is in a certain state
and ignore pairs that appear in different states.  Different memory
updating procedures can be created by choosing different states in
which to record.  Diagram (\ref{learning1}) illustrates the first
procedure.  The memory records only pairs that are adjacent in the
input.  The three lines show three different time steps and the two
windows show the words that the processor is trying to link.

\begin{sentence} \label{learning1}
\mbox{\psfig{file=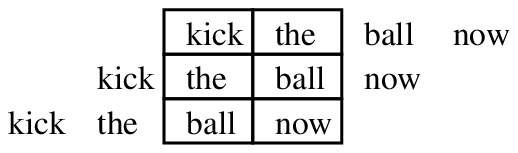}}
\end{sentence}

The basic data structures in the memory are words and word pairs with
their counts.  The same word may appear in the same window more than
once as the processor tries to link it with different words.  To keep
a consistent count, each word keeps track of how many times it was
seen in the left window and how many times it was seen in the right
window, in addition to how many times it actually appeared in the
text.  Then the lexical attraction of two words can be estimated as:

\begin{eqnarray*}
MI(x,y) & = & \log_2 \frac{P(x,y)}{P(x,*) P(*,y)} \\
& = & \log_2 \frac{n(x,y)/N}{(n(x,*)/N)(n(*,y)/N)}\\
& = & \log_2 \frac{n(x,y) N}{n(x,*) n(*,y)}
\end{eqnarray*}

\noindent where $MI$ is mutual information, $*$ is a wild-card matching
every word, $P(x,y)$ is the probability of seeing $x$ on the left
window and seeing $y$ on the right window, $n(x,y)$ is the count of
$(x,y)$, $N$ is the total number of observations made in both windows.

A significant percentage of syntactic relations are between adjacent
words.  Using the first procedure, the program can discover the
syntactic relations that are generally seen between adjacent words.
For example, it can relate determiners and adjectives to nouns and
learn collocations such as {\em ``The New York Times''}.
Diagram~(\ref{learning2}) illustrates how the program learns to relate a
determiner to its noun.

\begin{sentence} \label{learning2}
\mbox{\psfig{file=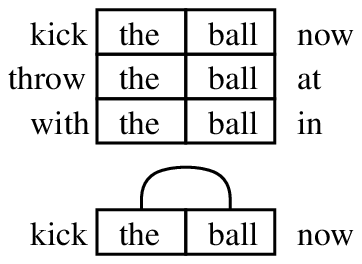}}
\end{sentence}

However, using this technique the program will never learn the
relationship between \word{kick} and \word{ball}.  A determiner almost
always intervenes.  There are other relations between words which are
practically never seen next to one another.  For example a
prepositional phrase modifying a transitive verb is always hidden
behind an object.

The second procedure for memory updating is to record all word pairs
encountered, no matter how far apart in each sentence.  Although this
procedure is guaranteed to see every related word pair at some point,
it also records a lot of unrelated pairs.  The next section shows that
even though this improves recall, the precision drops as expected.

Neither of the two procedures make use of the actual structure
identified by the processor.  In fact there is no feedback loop
between the processor and the memory, thus no real interdigitation of
learning and processing.  The memory gathers its information looking
at raw data and feeds it one way to the processor.

The third procedure is based on the feedback idea.  The memory only
records a subset of the pairs selected according to the structure
identified by the processor.  Before there is any structure, the
memory behaves as in the first procedure, recording only adjacent
pairs.  If AXB is a sequence of three words, then the pairs A-X and
X-B are recorded.  When two words are linked, they form a group.  In a
sense, they act like a single word.  If AX$\ldots$YB is a sequence of
words and X is linked to Y, the processor tries linking words in this
group to both A and B.  Because of the link crossing rule, words
between X and Y cannot be linked\footnote{Unless they are attracted
very strongly and are able to break the X-Y link.}.  So, in addition
to the adjacent pairs A-X and Y-B, the processor attempts to link A-Y
and X-B.  The memory records these pairs.

\begin{sentence} \label{learning3}
\mbox{\psfig{file=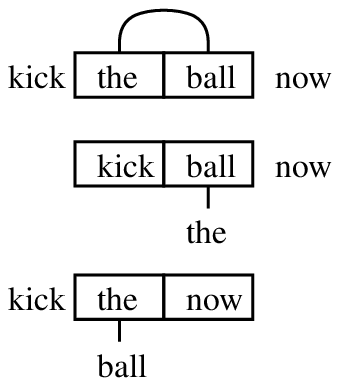}}
\hspace{1in}
\mbox{\psfig{file=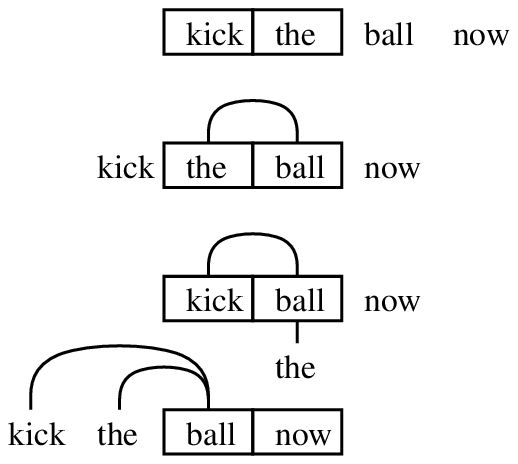}}
\end{sentence}

Diagram~(\ref{learning3}) shows how the third procedure leads to the
discovery of the relation between \word{kick} and \word{ball}.  In the
left figure, the program uses the rule given in the previous paragraph
and updates the pairs {\em kick-ball} and {\em the-now}.  The {\em
kick-ball} pair will be frequently reinforced and its high mutual
information will be detected.  The {\em the-now} pair will not be seen
together much more than expected, so its mutual information will stay
low.  The right figure illustrates how the long distance link is
formed once the correlation is identified.  The next section shows
that interdigitating learning and processing improves both precision
and recall.

\section{Results} \label{results}

This section presents results on the accuracy of the program in
finding relations between content-words.  I chose to base my
evaluation on content-word links because they are essential in the
extraction of meaning.  Content words are words that convey meaning
such as nouns, verbs, adjectives, and adverbs, as opposed to function
words, which convey syntactic structure, like prepositions and
conjunctions.  The mistakes in content-word links are significantly
more important than the mistakes in function-word links.  The two
sentences in (\ref{evaluation}) illustrate the difference.  In the
first sentence, a mistake would result in choosing the wrong subject
for flying.  In the second sentence, once the program has detected the
relation between money and education, which way the word
\word{for} links is less important.

\vspace*{20pt}
\begin{sentence} \label{evaluation}
\mylinkage{[[(I)()][(saw)()][(the)()][(mountains)()][(flying)()][(over)()][(New York)()]]}{[[3 4 0 (?)][0 4 1 (?)]]}

\vspace*{20pt}\hspace{50pt}
\mylinkage{[[(People)()][(want)()][(more)()][(money)()][(for)()][(education)()]]}{[[3 4 0 (?)][4 5 0 (?)][3 5 1 ()]]}
\end{sentence}

\begin{figure}[ht]
\centering
\mbox{\psfig{file=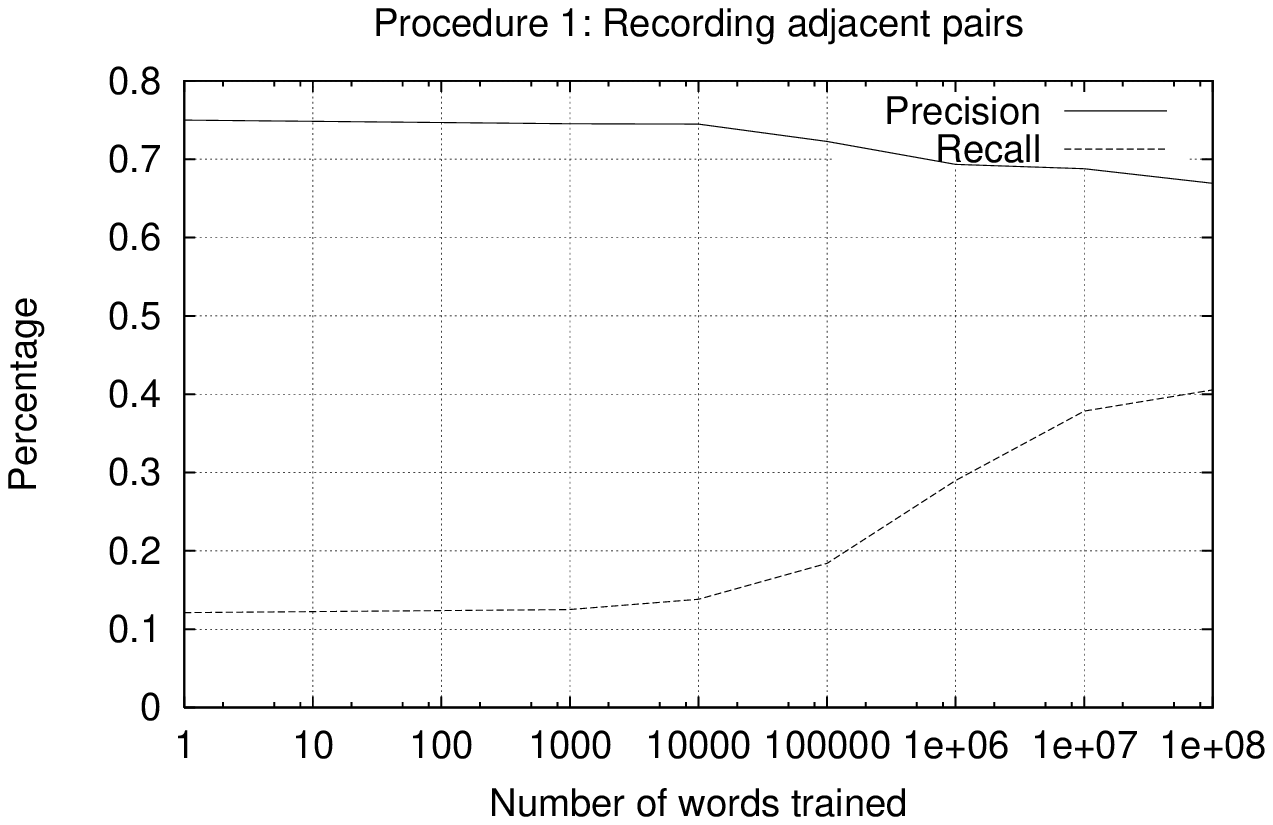,width=2.9in}
\psfig{file=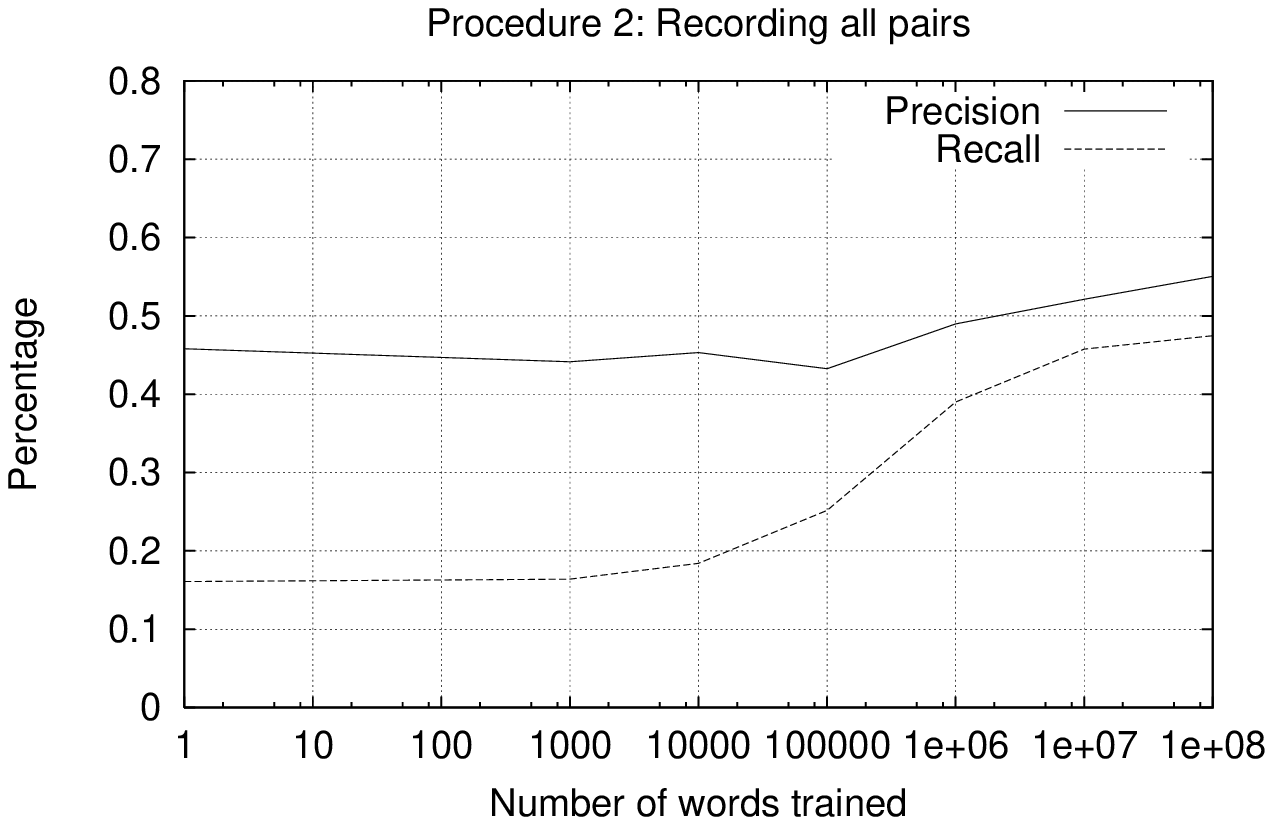,width=2.9in}} \\
\mbox{\psfig{file=figures/results.feedback.ps,width=2.9in}}
\caption{Content-word link accuracy results}
\label{accuracy}
\end{figure}

Figure~\ref{accuracy} shows the results of the three procedures
described in the previous section.  All three programs were trained
using up to 100 million words of Associated Press material.  After
training they were presented with 200 sentences of out-of-sample test
data and precision and recall were measured.  The 200 out of sample
sentences were hand parsed.  There were a total of 3152 words,
averaging 15.76 words per sentence, and 1287 content-word links in the
test set.  The choice of content-word links as an evaluation metric is
also significant here.  Most people agree on which content words are
related in a sentence, whereas even professional linguists argue about
how to link the function words.

There are two sources of mistakes for the program.  It can miss some
links because there is an unknown word in the sentence.  Or it can
make mistakes because of the failure of the processor or the learning
paradigm.  In order to isolate the two, the test sentences were
restricted to a vocabulary of 5,000 most frequent words, which account
for about 90\% of all the words seen in the corpus.

Accuracy is measured with precision and recall.  Recall is defined as
the percentage of content-word links present in the human parsed test
set that were recovered by the program.  Precision is defined as the
percentage of content-word links given by the program which were also
in the human parsed test set.

To help the reader judge the quality of these results, I will give
some numbers for comparison.  In the algorithm described above, the
processor consults the memory for the lexical attraction of two words
every time it is considering a new link.  When the program is modified
such that these numbers are supplied randomly, the precision is 8.9\%
and the recall is 5.4\%, which gives a lower bound.

The program relies on the assumption that syntactically related words
are also statistically correlated.  Of the 1287 pairs of words linked
by humans in the test set, 85.7\% of them actually had positive
lexical attraction, which gives an upper bound on recall.

\section{A critical evaluation} \label{critical}

In this section I present a qualitative analysis of the program's
shortcomings and suggest future work.  The mistakes of the program can
be traced back to three main reasons:

\begin{itemize}
\item There is no differentiation between different types of syntactic
relations.
\item The program does not represent or learn argument structures for
words. 
\item There is no mechanism for categorization and generalization.
\end{itemize}

\subsubsection*{Link types and second degree models}

Consider the following sentences:

\begin{sentence}
The architect worked on the building.
\end{sentence}
\begin{sentence}
The architect worked in the building.
\end{sentence}

The relation between \word{work} and \word{building} in the two
sentences are different.  This is marked by using different
prepositions.  In general, language can use word order, function
words, or morphology to represent different relations.

My program cannot represent this difference because it does not have
different link types and its independence assumptions do not permit
the representation of relations involving more than two words.  In the
examples above, the relation between two words is mediated by a third
word, the preposition.  The distribution of {\em in the building}, is
different from just {\em the building}.  A natural extension of my
approach would be to relax the independence assumptions one step and
to look at second-degree relationships, i.e. the modifiers of two
words should influence their lexical attraction and should be used to
mark the type of link between them.

\subsubsection*{Argument structure and using history}

Another related source of mistakes for the program is to link strongly
attracted words no matter how many other links they already have.
Because there is no representation of different link types, the
program cannot distinguish complements (mandatory arguments), from
adjuncts (optional arguments).  It can link four direct objects to a
verb, or have a determiner modify three nouns simultaneously.  After
it learns to represent different link types, the solution would be to
use the usage history of a word to learn its argument structure, and
use this information to constrain its relations.

This solution could also help the opposite problem.  The program
cannot link the words it has never seen together before.  Chomsky's
sentence is the ultimate example: {\em ``Colorless green ideas sleep
furiously.''}  The program would not be able to find any relations in
this sentence no matter how much training it goes through (unless it
comes across some linguistics textbooks).  Although in real data no
sentence is quite that bad, 15\% of the relations are between words
that do not have positive lexical attraction as pointed out in
Section~\ref{results}.  Learning from the usage history of words and
using this knowledge in less familiar situations is one solution to
this problem.

\subsubsection*{Categorization and generalization}

I stayed away from word categories in this work, partly because I
think they have done more harm than good in the past.  However, the
existence of word categories and their regulatory role in syntactic
constructions cannot be denied, although I believe we need a much
finer system of type hierarchy than the current subcategorization
frames and dictionary features allow.  

Learning about each word individually has many advantages.  However,
no matter how much data is used, some words will be seen few times.
One way to estimate the argument structure or the lexical attraction
information for rare words is to identify their category from the few
examples seen, and generalize the properties of this category to the
rare word.

In addition to providing a solution to low sample problem,
categorization of words is interesting in its own right.  A lot of
useful semantic information can be gathered by reading free text.  For
example, after reading ten years of Wall Street Journal, a program
could discover that there is a category corresponding to our concept
of politician, its members include the president, the prime-minister,
the governor, their frequent actions include visiting, meeting, giving
press releases etc.

There have been attempts to cluster words according to their usage,
see for example \cite{Pereira92a,Brown92,Lee97b}.  However their
success has been limited partly due to the lack of a good model to
identify syntactic relations in free text.  My approach is
particularly suitable for this line of work.

\section{Related work}

Most work on unsupervised language acquisition to date has used a
framework originally developed for finite state systems, i.e. (hidden)
Markov models and used in speech recognition.  The pillars of this
approach are two algorithms for training and processing with
probabilistic language models.  The Viterbi algorithm selects the most
probable analysis of a sentence given a model \cite{Viterbi67}.  The
Baum-Welch algorithm estimates the parameters of the model given a
sequence of training data \cite{Baum72}.  These algorithms are
generalized to work with probabilistic context free grammars in
addition to HMM's \cite{Baker79}.  The Baum-Welch is sometimes called
the {\em forward-backward} algorithm in the context of HMM's and the
{\em inside-outside} algorithm in the context of PCFG's.  For a
detailed description of these algorithms, see
\cite{Rabiner86,Lari90,Charniak93b}.

\begin{figure}[h]
\centering
\mbox{\psfig{file=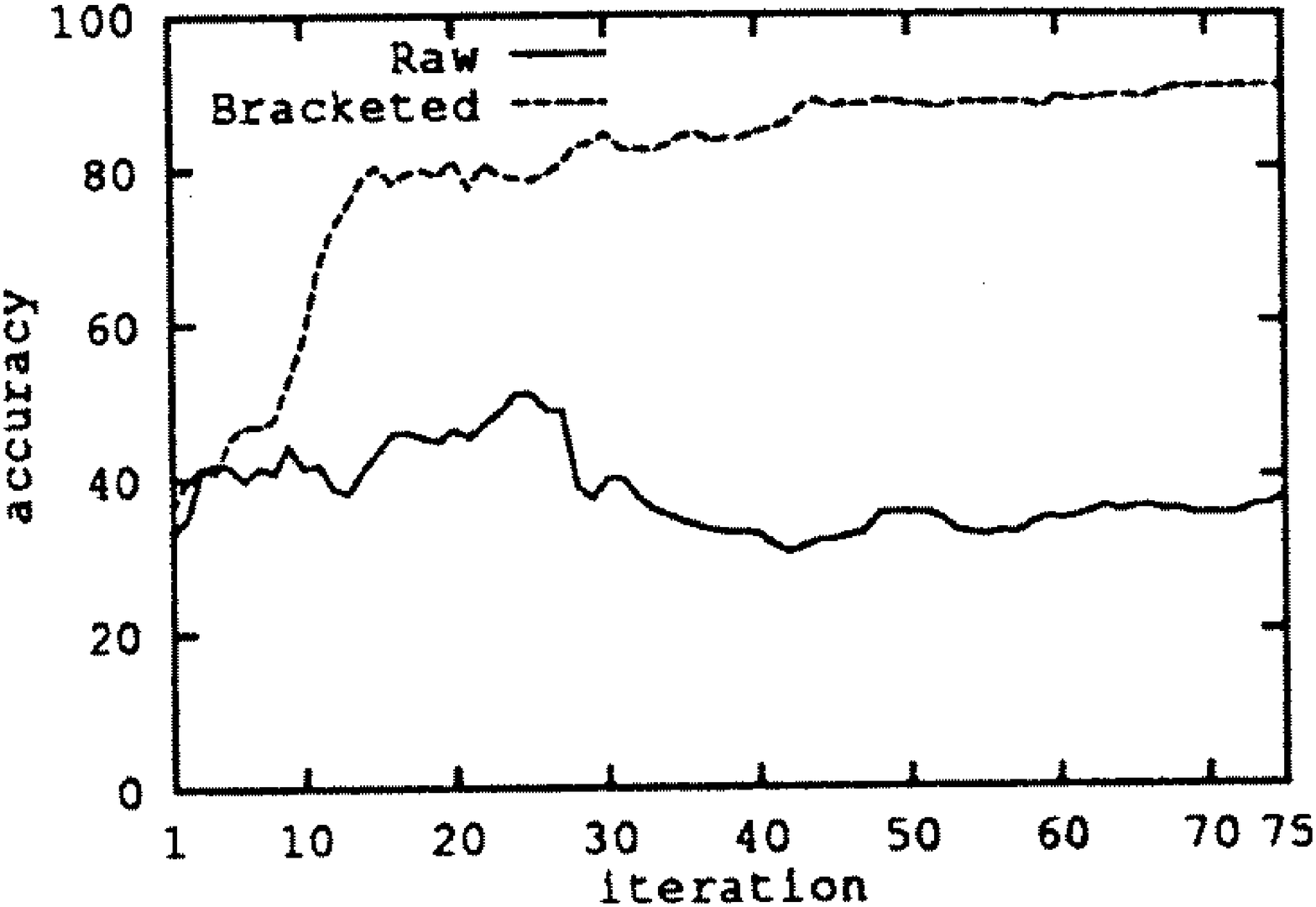,width=2.9in}
\psfig{file=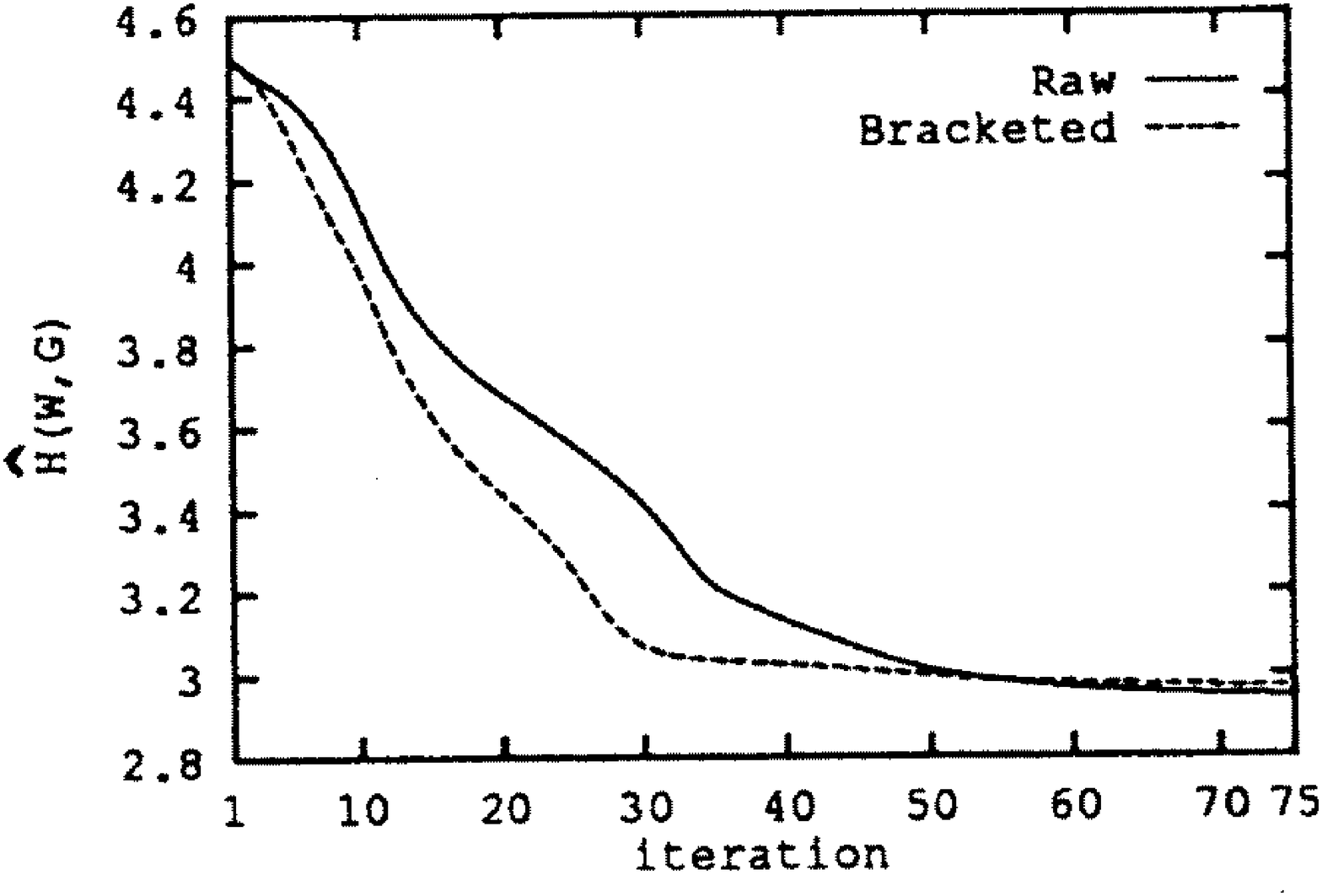,width=2.9in}}
\caption{No improvement in accuracy when trained with raw text.}
\label{pereira}
\end{figure}

The general approach is to define a space of context-free grammars and
improve the rule probabilities by training on a part-of-speech tagged
and sometimes bracketed corpus.  Different search spaces, starting
points, and training methods have been investigated by various
researchers.  Early work focused on optimizing the parameters for
hand-built grammars \cite{Jelinek85,Fujisaki89,Sharman90}.  Lari and
Young used the inside-outside algorithm for grammar induction using an
artificially generated language \cite{Lari90}.  Their algorithm is
only practical for small category sets and does not scale up to a
realistic grammar of natural language.  Carroll and Charniak used an
incremental approach where new rules are generated when existing rules
fail to parse a sentence \cite{Carroll92a,Carroll92b}.  Their method
was tested on small artificial languages and only worked when the
grammar space was fairly restricted.  Briscoe and Waegner started with
a partial initial grammar and achieved good results training on a
corpus of unbracketed text \cite{Briscoe92}.  Pereira and Schabes
started with all possible Chomsky normal form rules with a restricted
number of nonterminals and trained on the Air Travel Information
System spoken language corpus \cite{Pereira92}.  They achieved good
results training with the bracketed corpus but the program showed no
improvement in accuracy when trained with raw text.  Even though the
entropy improved, the bracketing accuracy stayed around 37\% for raw
text.  Figure~\ref{pereira} gives the accuracy and entropy results
from this work.  

In general these approaches fail on unsupervised acquisition because
of the large size of the search space, the complexity of the
estimation algorithm and the problem of local maxima.  Charniak
provides a detailed review of this work
\cite{Charniak93b}.  More recent work has focused on improving the
efficiency of the training methods
\cite{Stolcke94,Chen96b,deMarcken96a}.

\begin{figure}[h]
\centering
\fbox{\psfig{file=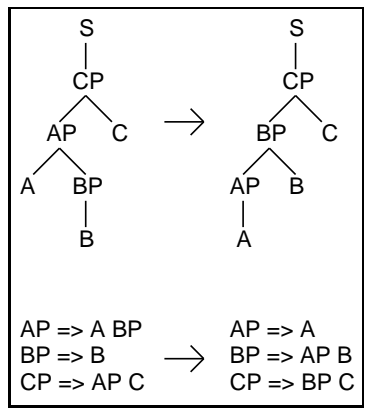,height=2in}}
\caption{Phrase structure representation makes it difficult to get out
of local maxima.}
\label{demarcken}
\end{figure}

de Marcken \cite{deMarcken95b} has an excellent critique on why the
current approaches to unsupervised language learning fail.  His main
observation is that the phrase structure representation makes it
difficult to get out of local maxima.  His other observation is that
early generalizations commit the learning programs to irrecoverable
mistakes.  Figure~\ref{demarcken} illustrates the impact of the
representation on the learning process.  Two different structures for
the string ABC are presented.  A modifies B in one of them and B
modifies A in the other.  If the learner has chosen the wrong one,
three nonterminals must have their most probable rules simultaneously
switched to fix the mistake.

My work contrasts with these approaches in two important respects.
First, most previous work uses part of speech information rather than
words themselves.  Recent work on high performance probabilistic
parsers confirm that detailed lexical information is needed for high
coverage accurate parsing \cite{Magerman95a,Collins96,Charniak97}.
Second, the standard formalism is focused on learning a probability
distribution over the possible tree structures.  My work simply
assumes a uniform distribution over all admissible trees and
concentrates on learning relationships between words instead.

\chapter{Contributions}
\label{contributions}

This work has been motivated by my desire to understand human language
learning ability and to build programs that can understand language.
Therefore the design decisions were given with a view towards
extraction of meaning.  Representing syntactic relations between words
directly is a consequence of having this goal.  The primitive
operation in the standard phrase-structure formalism is to group words
or phrases to form higher order constituents.  Meaningful relations
between words is an indirect outcome of the grouping process.  The
primitive operation in my work {\em is} finding meaningful relations
between words:

\vspace{20pt}

\begin{center}
\fbox{
\parbox{4.5in}{
\vspace*{10pt}\hspace{1pt}
\mylinkage{[[(The)()][(pilot)()][(saw)()][(the)()][(train)()][(flying)()][(over)()][(washington)()]]}{[[1 5 0 ()]]}

\vspace*{10pt}\hspace{1pt}
\mylinkage{[[(The)()][(driver)()][(saw)()][(the)()][(airplane)()][(flying)()][(over)()][(washington)()]]}{[[4 5 0 ()]]}
}}
\end{center}

\vspace{20pt}

The likelihood of two words being related is defined as {\em lexical
attraction}.  Knowledge of lexical attraction between words play an
important role in both language processing and acquisition.  I
developed a language program in which lexical attraction is the only
explicitly represented linguistic knowledge.  In contrast to other
work in language processing or acquisition, my program does not have a
grammar or a lexicon with word categories.  Chapter~\ref{theory}
formalizes lexical attraction within the context of information
theory:

\vspace{20pt}

\begin{center}
\fbox{\parbox{4.5in}{
\vspace*{10pt}
\mbox{
\mylinkage{[[(Northern)(1.48)][(Ireland)(14.65)]]}{[[0 1 0 (<)]]}\hspace{1.2in}
\mylinkage{[[(Northern)(12.60)][(Ireland)(3.53)]]}{[[0 1 0 (>)]]}\hspace{1.2in}
\mylinkage{[[(Northern)(12.60)][(Ireland)(14.65)]]}{[[0 1 0 (11.12)]]}
}
\vspace*{20pt}
}}
\end{center}

\vspace{20pt}

The program starts processing raw language input with no initial
knowledge.  It is able to discover more meaningful relations between
words as it processes more language.  The bootstrapping is achieved by
the interdigitation of learning and processing.  The processor uses
the regularities detected by the learner to impose structure on the
input.  This structure enables the learner to detect higher level
regularities which are difficult to see in raw input.
Chapter~\ref{learning} discusses the bootstrapping process:

\vspace{20pt}

\begin{center}
\fbox{\psfig{file=figures/bootstrap.ps}}
\end{center}

\vspace{20pt}

Starting with no knowledge and training on raw data, the program was
able to achieve 60\% precision and 50\% recall in finding relations
between content-words.  This is a significant result as previous work
in unsupervised language acquisition demonstrated little improvement
when started with zero knowledge.  

\vspace{20pt}

\begin{center}
\mbox{\psfig{file=figures/results.feedback.ps,width=4in}}
\end{center}

\vspace{20pt}

The key insights that differentiate my approach from others are:

\begin{itemize}
\item Training with words instead of parts of speech enables the
program to learn common but idiosyncratic usages of words.
\item Not committing to early generalizations prevent the program from
making irrecoverable mistakes.
\item Using a representation that makes the relevant features explicit
simplifies learning.  I believe what makes humans good learners is not
sophisticated learning algorithms but having the right representations.
\end{itemize}

This work has potential applications in semantic categorization and
information extraction.  More importantly it may shed light on how
humans are able to learn language from raw data and easily understand
syntactically ambiguous sentences.

\bibliography{nlp,book}

\begin{thebibliography}{}

\bibitem[\protect\citeauthoryear{Baker}{1979}]{Baker79}
Baker, J.
\newblock 1979.
\newblock Trainable grammars for speech recognition.
\newblock In {\em Speech communication papers presented at the 97th Meeting of
  the Acoustical Society},  547--550.

\bibitem[\protect\citeauthoryear{Baum}{1972}]{Baum72}
Baum, L.~E.
\newblock 1972.
\newblock An inequality and associated maximization technique in statistical
  estimation for probabilistic functions of markov processes.
\newblock {\em Inequalities} 3:1--8.

\bibitem[\protect\citeauthoryear{Beeferman, Berger, \&
  Lafferty}{1997}]{Beeferman97a}
Beeferman, D.; Berger, A.; and Lafferty, J.
\newblock 1997.
\newblock A model of lexical attraction and repulsion.
\newblock In {\em ACL/EACL '97}.

\bibitem[\protect\citeauthoryear{Briscoe \& Waegner}{1992}]{Briscoe92}
Briscoe, T., and Waegner, N.
\newblock 1992.
\newblock Robust stochastic parsing using the inside-outside algorithm.
\newblock In {\em AAAI '92 Workshop on Probabilistically-Based Natural Language
  Processing Techniques},  39--53.

\bibitem[\protect\citeauthoryear{Brown \& others}{1992}]{Brown92}
Brown, P.~F., et~al.
\newblock 1992.
\newblock Class-based n-gram models of natural language.
\newblock {\em Computational Linguistics} 18(4):467--479.

\bibitem[\protect\citeauthoryear{Carroll \& Charniak}{1992a}]{Carroll92a}
Carroll, G., and Charniak, E.
\newblock 1992a.
\newblock Learning probabilistic dependency grammars from labeled text.
\newblock In {\em Probabilistic Approaches to Natural Language, Papers from
  1992 AAAI Fall Symposium},  25--31.

\bibitem[\protect\citeauthoryear{Carroll \& Charniak}{1992b}]{Carroll92b}
Carroll, G., and Charniak, E.
\newblock 1992b.
\newblock Two experiments on learning probabilistic dependency grammars from
  corpora.
\newblock In {\em Workshop Notes, Statistically Based NLP Techniqies, AAAI},
  1--13.

\bibitem[\protect\citeauthoryear{Charniak}{1993}]{Charniak93b}
Charniak, E.
\newblock 1993.
\newblock {\em Statistical language learning}.
\newblock MIT Press.

\bibitem[\protect\citeauthoryear{Charniak}{1997}]{Charniak97}
Charniak, E.
\newblock 1997.
\newblock Statistical parsing with a context-free grammar and word statistics.
\newblock In {\em AAAI'97}.

\bibitem[\protect\citeauthoryear{Chen}{1996}]{Chen96b}
Chen, S.~F.
\newblock 1996.
\newblock {\em Building probabilistic models for natural language}.
\newblock Ph.D. Dissertation, Harvard University.

\bibitem[\protect\citeauthoryear{Chomsky}{1957}]{Chomsky57}
Chomsky, N.
\newblock 1957.
\newblock {\em Syntactic Structures}.
\newblock Mouton.

\bibitem[\protect\citeauthoryear{Chomsky}{1965}]{Chomsky65}
Chomsky, N.
\newblock 1965.
\newblock {\em Aspects of the theory of syntax}.
\newblock MIT Press.

\bibitem[\protect\citeauthoryear{Collins}{1996}]{Collins96}
Collins, M.~J.
\newblock 1996.
\newblock A new statistical parser based on bigram lexical dependencies.
\newblock In {\em Proceedings of the 34th Annual Meeting of the ACL}.

\bibitem[\protect\citeauthoryear{Cormen, Leiserson, \& Rivest}{1990}]{Cormen90}
Cormen, T.~H.; Leiserson, C.~E.; and Rivest, R.~L.
\newblock 1990.
\newblock {\em Introduction to Algorithms}.
\newblock MIT Press and McGraw-Hill.

\bibitem[\protect\citeauthoryear{Cover \& Thomas}{1991}]{Cover91}
Cover, T.~M., and Thomas, J.~A.
\newblock 1991.
\newblock {\em Elements of Information Theory}.
\newblock John Wiley and Sons, Inc.

\bibitem[\protect\citeauthoryear{de Marcken}{1995}]{deMarcken95b}
de~Marcken, C.~G.
\newblock 1995.
\newblock On the unsupervised acquisition of phrase-structure grammars.
\newblock In {\em Third Workshop on Very Large Corpora}.

\bibitem[\protect\citeauthoryear{de Marcken}{1996}]{deMarcken96a}
de~Marcken, C.~G.
\newblock 1996.
\newblock {\em Unsupervised language acquisition}.
\newblock Ph.D. Dissertation, MIT.

\bibitem[\protect\citeauthoryear{Fujisaki, Jelinek, \&
  others}{1989}]{Fujisaki89}
Fujisaki, T.; Jelinek, F.; et~al.
\newblock 1989.
\newblock A probabilistic parsing method for sentence disambiguation.
\newblock In {\em Proceedings of the 1st International Workshop on Parsing
  Technologies},  85--94.

\bibitem[\protect\citeauthoryear{Gaifman}{1965}]{Gaifman65}
Gaifman, H.
\newblock 1965.
\newblock Dependency systems and phrase-structure systems.
\newblock {\em Information and Control} 8:304--337.

\bibitem[\protect\citeauthoryear{Graham, Knuth, \& Patashnik}{1994}]{Graham94}
Graham, R.~L.; Knuth, D.~E.; and Patashnik, O.
\newblock 1994.
\newblock {\em Concrete Mathematics}.
\newblock Addison-Wesley, 2 edition.

\bibitem[\protect\citeauthoryear{Harary}{1969}]{Harary69}
Harary, F.
\newblock 1969.
\newblock {\em Graph Theory}.
\newblock Addison-Wesley.

\bibitem[\protect\citeauthoryear{Hudson}{1984}]{Hudson84}
Hudson, R.~A.
\newblock 1984.
\newblock {\em Word Grammar}.
\newblock B. Blackwell.

\bibitem[\protect\citeauthoryear{Jelinek}{1985}]{Jelinek85}
Jelinek, F.
\newblock 1985.
\newblock Markov source modeling of text generation.
\newblock In Skwirzinski, J.~K., ed., {\em The Impact of Processing Techniques
  on Communications}. Martinus Nijhoff.
\newblock  569--598.

\bibitem[\protect\citeauthoryear{Lari \& Young}{1990}]{Lari90}
Lari, K., and Young, S.
\newblock 1990.
\newblock The estimation of stochastic context-free grammars using the
  inside-outside algorithm.
\newblock {\em Computer Speech and Language} 4(1):35--56.

\bibitem[\protect\citeauthoryear{Lee}{1997}]{Lee97b}
Lee, L.
\newblock 1997.
\newblock {\em Similarity-based approaches to natural language processing}.
\newblock Ph.D. Dissertation, Harvard University.

\bibitem[\protect\citeauthoryear{Magerman}{1995}]{Magerman95a}
Magerman, D.~M.
\newblock 1995.
\newblock Statistical decision-tree models for parsing.
\newblock In {\em Proceedings of the 33rd Annual Meeting of the ACL}.

\bibitem[\protect\citeauthoryear{Mel'\v{c}uk}{1988}]{Melcuk88}
Mel'\v{c}uk, I.~A.
\newblock 1988.
\newblock {\em Dependency Syntax: Theory and Practice}.
\newblock SUNY.

\bibitem[\protect\citeauthoryear{Pearl}{1988}]{Pearl88}
Pearl, J.
\newblock 1988.
\newblock {\em Probabilistic Reasoning in Intelligent Systems: Networks of
  Plausible Inference}.
\newblock Morgan Kaufmann.

\bibitem[\protect\citeauthoryear{Pereira \& Schabes}{1992}]{Pereira92}
Pereira, F.~C., and Schabes, Y.
\newblock 1992.
\newblock Inside-outside reestimation from partially bracketed corpora.
\newblock In {\em Proceedings of the 30th Annual Meeting of the Association for
  Computational Linguist},  128--135.

\bibitem[\protect\citeauthoryear{Pereira \& Tishby}{1992}]{Pereira92a}
Pereira, F., and Tishby, N.
\newblock 1992.
\newblock Distributional similarity, phase transitions and hierarchical
  clustering.
\newblock In {\em Probabilistic Approaches to Natural Language, Papers from
  1992 AAAI Fall Symposium},  108--112.

\bibitem[\protect\citeauthoryear{Quirk \bgroup \em et al.\egroup
  }{1985}]{Quirk85}
Quirk, R.; Greenbaum, S.; Leech, G.; and Svartvik, J.
\newblock 1985.
\newblock {\em A Comprehensive Grammar of the English Language}.
\newblock Longman.

\bibitem[\protect\citeauthoryear{Rabiner \& Juang}{1986}]{Rabiner86}
Rabiner, L., and Juang, B.
\newblock 1986.
\newblock An introduction to hidden markov models.
\newblock {\em IEEE ASSP Magazine}  4--16.

\bibitem[\protect\citeauthoryear{Schank \& Colby}{1973}]{Schank73}
Schank, R.~C., and Colby, K.~M.
\newblock 1973.
\newblock {\em Computer Models of Thought and Language}.
\newblock Freeman.

\bibitem[\protect\citeauthoryear{Shannon}{1948}]{Shannon48}
Shannon, C.~E.
\newblock 1948.
\newblock A mathematical theory of communication.
\newblock {\em The Bell System Technical Journal} 27.

\bibitem[\protect\citeauthoryear{Shannon}{1951}]{Shannon51}
Shannon, C.~E.
\newblock 1951.
\newblock Prediction and entropy of printed english.
\newblock {\em The Bell System Technical Journal} 30:50--64.

\bibitem[\protect\citeauthoryear{Sharman, Jelinek, \& Mercer}{1990}]{Sharman90}
Sharman, R.; Jelinek, F.; and Mercer, R.
\newblock 1990.
\newblock Generating a grammar for statistical training.
\newblock In {\em Proceedings of the Third DARPA Speech and Natural Language
  Workshop},  267--274.

\bibitem[\protect\citeauthoryear{Sleator \& Temperley}{1991}]{Sleator91}
Sleator, D., and Temperley, D.
\newblock 1991.
\newblock Parsing english with a link grammar.
\newblock Technical Report CMU-CS-91-196, CMU.

\bibitem[\protect\citeauthoryear{Sleator \& Temperley}{1993}]{Sleator93}
Sleator, D., and Temperley, D.
\newblock 1993.
\newblock Parsing english with a link grammar.
\newblock In {\em Third international workshop on parsing technologies}.

\bibitem[\protect\citeauthoryear{Stolcke}{1994}]{Stolcke94}
Stolcke, A.
\newblock 1994.
\newblock {\em Bayesian learning of probabilistic language models}.
\newblock Ph.D. Dissertation, University of California at Berkeley.

\bibitem[\protect\citeauthoryear{Viterbi}{1967}]{Viterbi67}
Viterbi, A.~J.
\newblock 1967.
\newblock Error bounds for convolutional codes and an asymptotically optimal
  decoding algorithm.
\newblock {\em IEEE Transactions on Information Processing} 13:260--269.

\bibitem[\protect\citeauthoryear{Zipf}{1949}]{Zipf49}
Zipf, G.~K.
\newblock 1949.
\newblock {\em Human Behavior and the Principle of Least Effort}.
\newblock Addison-Wesley.

\end{thebibliography}
\bibliographystyle{aaai}

\end{document}